\documentclass[pdflatex,sn-mathphys-num]{sn-jnl}

\usepackage{graphicx}%
\usepackage{multirow}%
\usepackage{amsmath,amssymb,amsfonts}%
\usepackage{amsthm}%
\usepackage{mathrsfs}%
\usepackage[title]{appendix}%
\usepackage{xcolor}%
\usepackage{textcomp}%
\usepackage{manyfoot}%
\usepackage{booktabs}%
\usepackage{algorithm}%
\usepackage{algorithmicx}%
\usepackage{algpseudocode}%
\usepackage{listings}%
\usepackage{subcaption}

\theoremstyle{thmstylethree}%
\usepackage{graphicx,afterpage}
\usepackage{bm}\usepackage{soul}
\usepackage{empheq}\usepackage{mathtools}
\usepackage{graphicx,afterpage,cancel}

\numberwithin{equation}{section}
\numberwithin{figure}{section}
\def\theequation{\arabic{section}.\arabic{equation}}
\newcommand\tabcaption{\def\@captype{table}\caption}
\def\thefigure{\arabic{section}.\arabic{figure}}

\newtheorem{example}{Example}[section] 
\renewcommand{\theexample}{\thesection.\arabic{example}} 



\renewcommand{\abstractname}{Summary}
\definecolor{orange}{RGB}{255,127,0}
\definecolor{gray}{RGB}{128,128,128}

\afterpage{\clearpage}

\title{Bridging Idealized and Operational Models: An Explainable AI Framework for Earth System Emulators}

\author[1]{\fnm{Pouria} \sur{Behnoudfar}}\email{behnoudfar@wisc.edu}

\author[1]{\fnm{Charlotte} \sur{Moser}}\email{crmoser2@wisc.edu}
\author[2]{\fnm{Marc} \sur{Bocquet}}\email{marc.bocquet@enpc.fr}

\author[2]{\fnm{Sibo} \sur{Cheng}}\email{sibo.cheng@enpc.fr}

\author[1]{\fnm{Nan} \sur{Chen$^*$}}\email{chennan@math.wisc.edu}

\affil[1]{\orgdiv{Department of Mathematics}, \orgname{University of Wisconsin-Madison}, \orgaddress{ \state{Wisconsin}, \country{USA}}}

\affil[2]{\orgdiv{CEREA}, \orgname{ ENPC, EDF R\&D, Institut Polytechnique de Paris, Île-de-France}, \orgaddress{ \country{France}}}

\begin{document}

\abstract{
Computer models are indispensable tools for understanding the Earth system. While high-resolution operational models have achieved many successes, they exhibit persistent biases, particularly in simulating extreme events and statistical distributions. In contrast, coarse-grained idealized models isolate fundamental processes and can be precisely calibrated to excel in characterizing specific dynamical and statistical features. However, different models remain siloed by disciplinary boundaries. By leveraging the complementary strengths of models of varying complexity, we develop an explainable AI framework for Earth system emulators. It bridges the model hierarchy through a reconfigured latent data assimilation technique, uniquely suited to exploit the sparse output from the idealized models. The resulting bridging model inherits the high resolution and comprehensive variables of operational models while achieving global accuracy enhancements through targeted improvements from idealized models. Crucially, the mechanism of AI provides a clear rationale for these advancements, moving beyond black-box correction to physically insightful understanding in a computationally efficient framework that enables effective physics-assisted digital twins and uncertainty quantification. We demonstrate its power by significantly correcting biases in CMIP6 simulations of El Ni\~no spatiotemporal patterns, leveraging statistically accurate idealized models. This work also highlights the importance of pushing idealized model development and advancing communication between modeling communities.}

\keywords{Explainable AI, Latent Data Assimilation, Conceptual Models, Earth System Modeling, Digital Twins, El Ni\~no}
\maketitle
\tableofcontents
\section{Introduction}

Computer models are fundamental tools for understanding and predicting complex dynamical systems across Earth science and other disciplines. Accurate model simulations are crucial not only for advancing scientific knowledge but also for addressing critical societal challenges, such as early warning of extreme events, managing environmental risks, and informing policy-making~\cite{bonan2018climate, calvin2018integrated, carbone2014managing}.

Operational models, especially high-resolution general circulation models (GCMs), are among the most commonly used computational tools to understand the Earth system~\cite{randall2000general, wilby1997downscaling}. These models encompass numerous known relevant processes across various spatiotemporal scales in the atmosphere, ocean, land, and ice, aiming to provide a detailed representation of the Earth system. Despite achieving many successes, these high-resolution operational models often suffer from persistent biases, resulting in non-negligible errors in simulating the detailed dynamics and statistics of major climate phenomena. For example, many operational models still have difficulties in fully capturing key features of the moisture dynamics of the Madden-Julian Oscillation (MJO)~\cite{lin2024assessment, le2021underestimated} and the diversity and extreme statistics of the El Ni\~no-Southern Oscillation (ENSO)~\cite{guilyardi2009understanding, guilyardi2020enso}. On the opposite end of the spectrum, idealized models, including conceptual and simple intermediate-coupled models, isolate fundamental dynamical processes and are coarse-grained by design~\cite{eby2013historical, saatsi2013idealized}. These low-dimensional models can be precisely calibrated to excel in characterizing specific large-scale features, where operational models fail. The computational efficiency and dynamical insight provided by these simple models have led to their renewed attention across various Earth system modeling communities~\cite{reed2025idealized, vialard2025nino, kumar2000analysis}.

Despite the complementary strengths, the models belonging to different classes remain largely siloed by disciplinary boundaries~\cite{held2005gap, palmer2016personal, stocker2011introduction}. While idealized models are expected to guide the improvement of operational models, directly revising the sophisticated operational models based on the qualitative insights from the idealized models is notoriously challenging, as adjustments to one component often introduce unintended consequences or numerical instabilities in other elements~\cite{mauritsen2012tuning, schmidt2015practical}. Conversely, the primary use of operational models for idealized models is limited to the performance evaluation of these simpler models~\cite{levine2017simple, bony2015clouds}. Therefore, while each model type has value, they are not leveraged synergistically. Addressing this gap is a central challenge in contemporary Earth system modeling~\cite{schneider2017earth, vialard2025nino}, which creates a critical need for developing a practical framework that can systematically translate the statistical accuracy and dynamical insights of simple models into direct and stable corrections for high-resolution operational models.

Machine learning (ML) has emerged as a powerful tool in Earth science, notably for creating efficient data-driven emulators~\cite{rasp2018deep}. However, a key criticism of many ML models is their black-box nature~\cite{samek2019explainable}. These standard ML models may interpolate training data well but lack physical interpretability. Therefore, they often fail to provide trustworthy insights for events outside their training distribution~\cite{reichstein2019deep}. This limitation is particularly problematic for improving physical models, as simply embedding an ML component does not elucidate the physical reasons for a model's biases~\cite{huntingford2019machine}. While the field of explainable AI (XAI) aims to address interpretability, XAI techniques often face a trade-off between accuracy and transparency~\cite{rudin2019stop}. In addition, most existing XAI efforts focus on interpreting stand-alone AI models~\cite{samek2019towards}.

In this paper, we develop an XAI framework to bridge the divide between idealized and operational models. Fundamentally different from conventional XAI, our framework is designed to integrate and correct physics-based models across the complexity hierarchy. Its centerpiece is a reconfigured latent space data assimilation technique, uniquely suited to exploit the sparse output of coarse-grained models. This enables the direct assimilation of targeted dynamical and statistical improvements from idealized models into high-resolution operational models. Crucially, the process is inherently explainable. Rather than acting as a black-box surrogate, the AI's core mechanism is to provide a transparent bridge between models. This design reveals a clear rationale for why each part of the resulting bridging model is improved, leading to a physically insightful understanding. The computationally efficient bridging model enables the generation of massive, high-quality datasets for quantifying the uncertainty of the Earth system simulations and analyzing extreme events. It also facilitates rapid sensitivity tests by computing responses in the idealized model and then assimilating them into the full field. Ultimately, this creates a new approach for designing digital twins that efficiently and robustly test what-if scenarios. More broadly, this framework highlights the crucial importance of advancing idealized model development and promoting communication among modeling communities.

We demonstrate our framework by enhancing the simulation of ENSO complexity, which encompasses the diverse spatial patterns, amplitudes, and temporal evolutions of the ENSO phenomenon, in CMIP6 models~\cite{beobide2021uncertainty}. By leveraging simpler yet statistically accurate conceptual and intermediate-coupled models, we significantly correct biases in the state-of-the-art operational systems, enhancing the representation of ENSO dynamics and statistics.

\section{Methods}
Operational and idealized models offer complementary yet isolated strengths. High-resolution operational models provide a comprehensive and physically detailed representation of the Earth system. However, they are computationally expensive and often exhibit persistent biases in key dynamics and statistics. In contrast, idealized models are computationally efficient. They can be precisely calibrated to excel in capturing specific large-scale processes and statistical features, but they achieve this by design through coarse-graining and isolating only fundamental dynamics. The central challenge, therefore, lies in the lack of a systematic mechanism to transfer the targeted accuracy and insight from idealized models into the complex and high-resolution framework of operational models, thereby correcting their biases without inducing instability.

\begin{figure}[H]
    \centering
    \includegraphics[width=.9\linewidth]{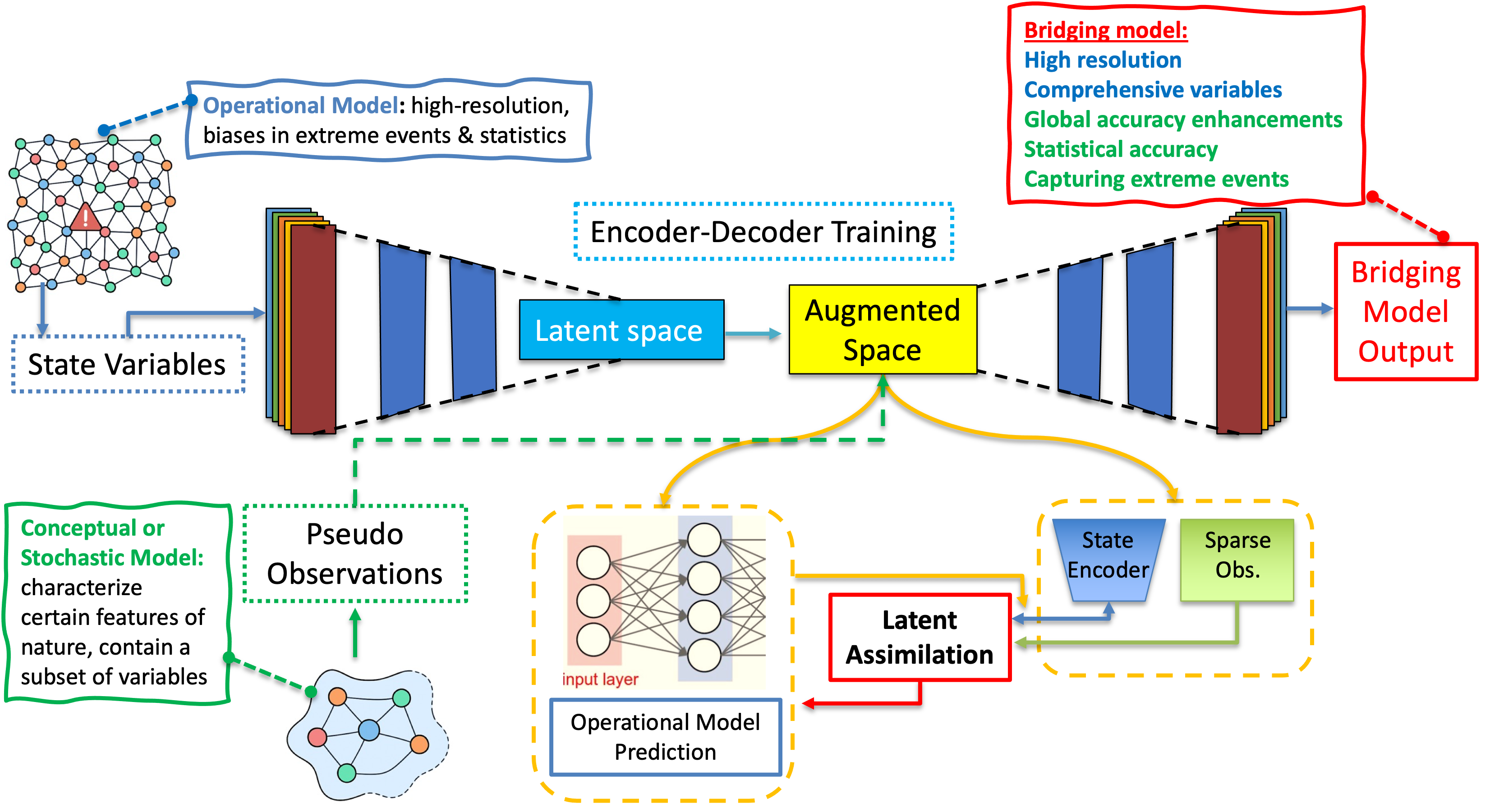}
    \caption{Overview of developing the bridging model from XAI.  }
    \label{fig:overview}
\end{figure}

The goal of our XAI framework is to systematically bridge the divide between idealized and operational models, creating a bridging model that leverages their complementary strengths. An overview of the framework is presented in Figure~\ref{fig:overview}. It consists of four interconnected components, each serving a distinct purpose in the model integration procedure.
\begin{itemize}
\item Physically-augmented latent space. An autoencoder creates the foundational representation by mapping the high-dimensional state variables of the operational model into a compressed latent space. Crucially, this latent space is then enriched with the observed physical quantities, forming an augmented latent space. This step, which is fundamentally distinguishable from standard latent space data assimilation approaches, ensures that the complex model state is not only reduced for computational efficiency but is also physically grounded and directly connected to the outputs of idealized models, serving as ``pseudo-observations'' (see below). This physically-augmented latent space plays a crucial role in advancing skillful data assimilation that bridges models with different complexities.
\item Data-driven short-term model forecasts in the augmented latent space. A data-driven forecast model is trained to emulate the short-term dynamical evolution of the operational model within this augmented latent space. This process provides the background forecast for the data assimilation cycle, efficiently propagating the state from the operational model forward to be corrected by the idealized models.
\item Model-bridging data assimilation. A reconfigured data assimilation process performs the core bridging function. At this stage, the forecasted state is updated by assimilating ``pseudo-observations'' provided by the idealized models. These models contribute targeted dynamical and statistical insights that are well-represented within the augmented latent space. A reconfigured latent space data assimilation algorithm is designed to be uniquely suited for exploiting the sparse yet physically meaningful output from coarse-grained idealized models, thereby directly injecting their accuracy into operational surrogates.
\item Correcting structural bias via curriculum learning. While data assimilation corrects the forecast state using idealized models, the initial latent representation itself is learned from the biased operational model, leaving certain structural errors uncorrected. To mitigate such biases, we employ a curriculum learning strategy. By gradually blending limited reanalysis data into the training process, we correct the foundational representation itself, resulting in a more accurate and observationally-constrained latent space for all subsequent steps.
\end{itemize}
This entire procedure facilitates the direct and stable assimilation of improvements from idealized models into operational models. It is inherently explainable, as the impact of each idealized model correction is traceable through the augmented latent space, providing a clear physical rationale for the resulting enhancements.


\subsection{Latent space representation and physical augmentation}

A central challenge in data assimilation for high-dimensional systems is the prohibitive computational cost. While standard autoencoders can reduce dimensionality, their latent spaces often lack direct physical interpretability, limiting their utility for integrating physics-based insights~\cite{peyron2021latent, cheng2023generalised}. To overcome this difficulty, we introduce a key innovation: a physically augmented latent space. We begin with a standard unsupervised autoencoder but fundamentally reconfigure its objective. The core idea is to enrich the compressed latent vector with physically meaningful observational data, facilitating the communication between model states and observations. Formally, a normalized input tensor $\mathbf{X}_k \in \mathbb{R}^{c \times h \times w}$, containing $c$ physical variables over the spatial grid points $h \times w$ at time $k$, is mapped by an encoder $\mathcal{E}(\cdot) $ to a create a latent representation. This latent representation is then augmented with a vector of sparse physical observations $\mathbf{y}_k$, yielding our augmented state vector:
\begin{equation}
    \mathbf{x}_k = \begin{bmatrix} \mathcal{E}(  \mathbf{X}_k) \\  \mathbf{y}_k \end{bmatrix} \in \mathbb{R}^{n_l+n_o}.
\end{equation}
Our latent space thus directly includes information from both the state and the newly available observations, making it different from existing latent-space data assimilation methods~\cite{cheng2023machine}.

To train this model, we define a composite loss function that jointly optimizes two objectives: minimizing the mean squared error (MSE) between the input and the reconstructed fields, and maximizing the cross-correlation between the latent space and the observables. They aim to preserve the high-fidelity spatial information of the operational model and ensure that the latent code is statistically correlated with key physical quantities, respectively. The loss function reads:
\begin{subequations}
\begin{align}
    \mathcal{L} &\;=\; \mathcal{L}_{\text{recon}} + \lambda \cdot \mathcal{L}_{\text{corr}},\\
    \mathcal{L}_{\text{recon}} &\;=\; \frac{1}{c h w} \sum_{l=1}^{c} \sum_{i=1}^{h} \sum_{j=1}^{w} \left( \mathbf{X}_{k,l,i,j} - \mathbf{\hat{X}}_{k,l,i,j} \right)^2,\\
    \mathcal{L}_{\text{corr}}  &\;=\; - \frac{1}{n_o n_l} \sum_{i=1}^{n_o} \sum_{j=1}^{n_l} C_{ij},
\end{align}
\end{subequations}
where $ \hat{\mathbf{X}}_k \in \mathbb{R}^{c \times h \times w}$ denotes the reconstructed field output by the decoder. 
The correlation matrix $\mathbf{C}$ is determined as:
\begin{subequations}
    \begin{align}
\tilde{\mathbf{y}}_k &= \mathbf{y}_k - \frac{1}{n_o}\sum_{i=1}^{n_o} \mathbf{y}_{k,i},\\
\tilde{\mathbf{x}}_k &= \mathcal{E}(\mathbf{X}_k) - \frac{1}{n_l}\sum_{i=1}^{n_l} \mathcal{E}(\mathbf{X}_k)_i,\\
\mathbf{C} &= \frac{\tilde{\mathbf{y}}_k^\top \tilde{\mathbf{x}}_k}{\|\tilde{\mathbf{y}}_k\| \, \|\tilde{\mathbf{x}}_k\| + \varepsilon},
    \end{align}
\end{subequations}
where $\mathbf{C} \in \mathbb{R}^{n_l \times n_o}$, $\varepsilon$ is a small constant added for numerical stability, and $\lambda$ is a hyperparameter controlling the trade-off between reconstruction and physical alignment.

This dual-objective approach ensures the autoencoder not only preserves the essential spatial structures of the operational model but also forces an alignment between the latent space and key physical features. The result is a highly explainable representation specifically designed for downstream data assimilation. 

In the Supplementary Information, the role of $\lambda$ is examined, providing quantitative justification for the use of the augmented latent space and the necessity of incorporating the maximization of the cross-correlation between the latent space and the observed physical quantities as a second training objective.

\subsection{Data-driven short-term model forecasts in the augmented latent space}
With the system state now represented in the physically-augmented latent space, the next step is to forecast its short-term evolution. This forecast provides the essential background state for the subsequent data assimilation cycle. We model this temporal dynamics using a recurrent neural network based on Long Short-Term Memory (LSTM) units~\cite{sherstinsky2020fundamentals}. The LSTM network $\mathcal{F}_{\text{LSTM}}$ is trained to predict the next augmented state $\mathbf{x}_{t+1}$ given a sequence of previous states $\mathbf{x}_1,\,\cdots, \mathbf{x}_t$. The network updates its internal states through the following gating mechanisms~\cite{yu2019review}:
\begin{align*}
\mathbf{i}_t &= \sigma\left( \mathbf{W}_i \mathbf{z}_t + \mathbf{U}_i \mathbf{h}_{t-1} + \mathbf{b}_i \right), \quad 
&&\mathbf{h}_t = \mathbf{o}_t \odot \tanh(\mathbf{c}_t),\\
\mathbf{g}_t &= \tanh\left( \mathbf{W}_g \mathbf{z}_t + \mathbf{U}_g \mathbf{h}_{t-1} + \mathbf{b}_g \right), \quad
&&\mathbf{c}_t = \mathbf{f}_t \odot \mathbf{c}_{t-1} + \mathbf{i}_t \odot \mathbf{g}_t,\\ \mathbf{o}_t &= \sigma\left( \mathbf{W}_o \mathbf{z}_t + \mathbf{U}_o \mathbf{h}_{t-1} + \mathbf{b}_o \right),\quad
&&\mathbf{f}_t = \sigma\left( \mathbf{W}_f \mathbf{z}_t + \mathbf{U}_f \mathbf{h}_{t-1} + \mathbf{b}_f \right),
\end{align*}
where \( \mathbf{i}_t \), \( \mathbf{h}_t \), and \( \mathbf{o}_t \) are the input, hidden, and output states, respectively. \( \mathbf{c}_t \) is the cell state and \( \odot \) denotes element-wise multiplication. The weight matrices \( \mathbf{W}_{(\cdot)}\) and \(\mathbf{U}_{(\cdot)} \), and biases \( \mathbf{b}_{(\cdot)} \), are learnable parameters. \( \sigma(\cdot) \) denotes the activation function. We employ the ELU activation function for input and hidden layers, while a $\tanh$ output layer bounds predictions to $[-1,1]$.

The model is trained to minimize the mean squared error between the forecast and the true future (latent) states:
\begin{align}
  \mathbf{x}^f_{t+1} &= \mathcal{F}_{\text{LSTM}}\left( \mathbf{x}_1,\,\cdots, \mathbf{x}_t\right), \\
\mathcal{L}_{\text{MSE}} &=  \left\|  \mathbf{x}^f_{t+1} - \mathbf{x}_{t+1} \right\|.
\end{align}
Our implementation utilizes a two-layer LSTM architecture, where the size of the hidden state matches the dimension of the augmented latent vector. A held-out validation set (5\% of the data) is used for early stopping to prevent overfitting. This trained forecast model efficiently propagates the operational model's state within the structured latent space. The resulting forecast now serves as the prior, setting the stage for the final and most crucial step: its principled correction using insights from idealized models.

\subsection{Bridging via assimilation of idealized model output}

One of the key steps of our framework leverages data assimilation not to incorporate real-world observations, but to systematically inject the targeted insights of idealized models into the operational forecast. This reconfigures data assimilation from a tool for state estimation into a powerful mechanism for integrating model hierarchies.

The process begins with an ensemble forecast from the previous step: \\$\mathbf{x}^{f,(i)}_{t+1} = \mathcal{F}_{\text{LSTM}}\left(\mathbf{x}_1^{a,(i)},\,\cdots, \mathbf{x}_t^{a,(i)} \right)$  for $i=1,\cdots,N$. These forecasts are updated using a stochastic ensemble Kalman filter (EnKF) update:
\begin{subequations}
\begin{align}
    \mathbf{x}^{a,(i)}_{t+1} &= \mathbf{x}^{f,(i)}_{t+1} +\mathbf{K}(\mathbf{y}^{(i)}_{t+1} - \mathbf{H}\mathbf{x}^{f,(i)}_{t+1}),\\
      \mathbf{K} &= \mathbf{P}_{\text{xb}} \mathbf{H}^\top
    (\mathbf{H}\mathbf{P}_{\text{xb}}\mathbf{H}^\top + \mathbf{R})^{-1},
      \end{align}
\end{subequations}
where \(\mathbf{P}_{\text{xb}}\) is the background error covariance and \(\mathbf{R}\) is the observation error covariance. The details can be found in the Supplementary Information.

The critical innovation lies in the definition of the ``observation''. Here, $\mathbf{y}_{t+1}$ is not a real-world measurement but a pseudo-observation generated by an idealized model. The observation operator $\mathbf{H}$ is correspondingly defined to select only the last $n_o$ entries of the augmented state vector (the physical observables):
\begin{equation}
\mathbf{H} =
\begin{bmatrix}
\mathbf{0}_{ n_o \times n_l} & \mathbf{I}_{ n_o \times n_o}
\end{bmatrix}.
\end{equation}
This setup is what enables the bridge. Because the autoencoder's loss function maximized the cross-covariance between the latent variables $\mathcal{E}({\mathbf{X}})$ and the physical observables $\mathbf{y}$, the Kalman gain $\mathbf{K}$ computes corrections that propagate coherently through the entire augmented state. A correction to the physical observable $\mathbf{y}$ induces a physically consistent correction in the latent representation of the high-resolution field $\mathcal{E}({\mathbf{X}})$.

Therefore, this step does not merely nudge the model. It translates the dynamical and statistical accuracy of the idealized model directly into a stable, physically reasoned correction for the operational model, completing the explainable bridge across the model hierarchy.

\subsection{Correcting structural bias via curriculum learning}

A key challenge in building a skillful bridging model is to integrate not only the high-fidelity output from idealized models but also limited reanalysis data. To ensure our bridging model is not solely bound by the biases of the operational models yet can leverage the limited reanalysis data, we incorporate a curriculum learning strategy~\cite{wang2021survey} during the training of both the autoencoder and the LSTM forecast model.

\begin{figure}[H]
    \centering
    \includegraphics[width=.9\linewidth]{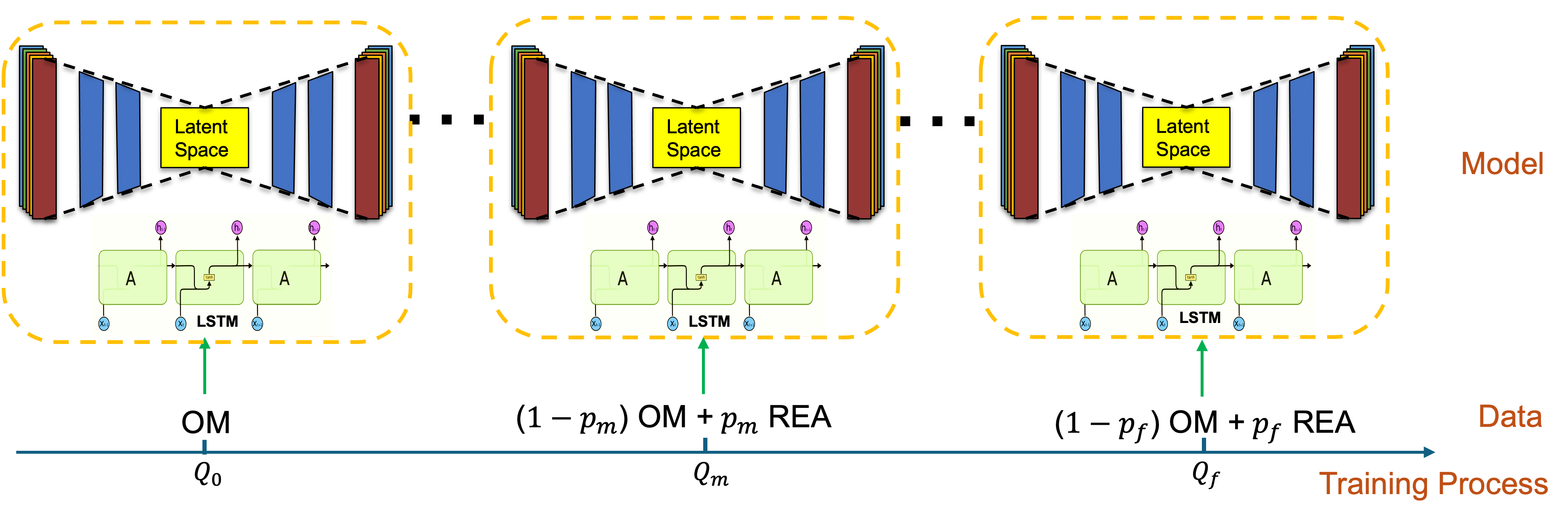}
    \caption{Schematic of the curriculum learning strategy for training the autoencoder and forecast models. The training data is gradually transitioned from purely operational model (OM) outputs to a blend of OM and reanalysis (REA) data. }
    \label{fig:Q_learn}
\end{figure}

This strategy gradually introduces reanalysis data into the training process. We define a curriculum schedule where the probability $p_e$ of including observational data at training epoch $e$ increases linearly from an initial value $p_0$ to a maximum $p_{\max}$:
\begin{equation}
    p_e = \min\!\left(p_{\max}, \; p_0 + \frac{e}{e_f} \big(p_{\max} - p_0 \big)\right),
\end{equation}
where $e_f$ is the total number of epochs. For our experiments, we set $p_0 = 0$ and $p_{\max} = 0.6$. Thus, the model begins training exclusively on the operational model data and is progressively exposed to more observational data, reaching a maximum inclusion probability of 60\% by the final epoch (see Figure~\ref{fig:Q_learn}).

It is worth highlighting that training the forecast model directly on limited reanalysis data would introduce substantial error due to overfitting. Therefore, we use the operational model as the primary training basis, with reanalysis data serving to refine its performance. This strategy ensures computational stability, prevents overfitting to sparse observations, and mitigates the operational model's inherent biases. The result is a robust and generalizable bridging model anchored in both simulated and real-world physics.


\section{Results}
We demonstrate the power of our XAI framework by substantially correcting persistent biases in the simulation of ENSO and its spatiotemporal complexity within state-of-the-art operational models, using targeted insights from statistically accurate idealized models.

ENSO is a dominant mode of interannual climate variability in the equatorial Pacific, with far-reaching impacts on global weather patterns~\cite{mcphaden2006enso, ropelewski1987global, ashok2007nino}. ENSO alternates between a warm phase, El Ni\~no, and a cool phase, La Ni\~na, which are identified by positive and negative sea surface temperature (SST) anomalies, respectively. Recent studies have called attention to ENSO complexity and diversity, which characterizes how ENSO events can vary across cycles~\cite{capotondi2015understanding, timmermann2018nino}. Specifically, ENSO diversity refers to the existence of two types of ENSO events distinguished by the location of their temperature anomaly centers: Central Pacific (CP) and Eastern Pacific (EP) events~\cite{ashok2007nino, kao2009contrasting, kim2012statistical}. More broadly, ENSO complexity encompasses variations in the ENSO cycle by not only spatial pattern, but also intensity and temporal evolution~\cite{chen2008nino, jin2008current, fang2015cloud, sohn2016strength, santoso2019dynamics}.

Despite significant improvements in the operational models, most of the Coupled Model Intercomparison Project Phase 6 (CMIP6) models~\cite{eyring2016overview} have difficulties in reproducing the realistic features of the two types of El Ni\~nos, let alone capturing the ENSO complexity in spatiotemporal patterns~\cite{zhang2024consistent, geng2025enhanced}. Among the CMIP6 models, the Community Earth System Model 2 (CESM2)~\cite{danabasoglu2020community} and the Geophysical Fluid Dynamics Laboratory Coupled Physical Model 4 (GFDL CM4)~\cite{held2019structure} are the most widely used models for understanding and predicting ENSO.

\subsection{Setup}
\subsubsection{Reanalysis data}
We employ the NCEP Global Ocean Data Assimilation System (GODAS)~\cite{behringer2004evaluation} as our reanalysis dataset, using monthly data from 1981 to 2023. This dataset serves as the benchmark for evaluating both the operational and bridging models and is incorporated into the curriculum learning strategy.

\subsubsection{Definitions of Ni\~no regimes and different types of the ENSO events}
The Ni\~no 3, Ni\~no 3.4, and Ni\~no 4 indices are defined as the sea surface temperature (SST) anomalies averaged over the tropical Pacific regions $150^{\circ}$W--$90^{\circ}$W, $170^{\circ}$W--$120^{\circ}$W, and $160^{\circ}$E--$150^{\circ}$W, respectively, each within the latitude band $5^{\circ}$S to $5^{\circ}$N. Following~\cite{kug2009two}, El Ni\~no and La Ni\~na events are identified based on these indices during boreal winter (December--February). An EP El Ni\~no is characterized by SST anomalies in the Ni\~no 3 region exceeding $0.5^{\circ}$C and larger than those in the Ni\~no 4 region. Events with a maximum Ni\~no 3 SST anomaly above $2.5^{\circ}$C (from April to the following March) are classified as extreme EP El Ni\~no~\cite{wang2019historical}. Conversely, a CP El Ni\~no occurs when anomalies in the Ni\~no 4 region exceed $0.5^{\circ}$C and are larger than those in the Ni\~no 3 region. La Ni\~na events are defined analogously by anomalies below $-0.5^{\circ}$C and are categorized as CP or EP based on the location of the peak cooling. Finally, events persisting for two or more consecutive years are classified as multi-year El Ni\~no or La Ni\~na.

\subsubsection{Operational models}

For the results presented in the main text, we utilize historical CESM2 outputs from 1850 to 2014, focusing on the equatorial Pacific domain spanning longitudes from $140^\circ\mathrm{E}$ to $80^\circ\mathrm{W}$ ($140^\circ$ to $280^\circ$) and latitudes from $-15^\circ$ to $15^\circ$, with 1-degree spatial resolution. The primary anomaly variables considered are zonal wind stress ($\tau_x$), sea surface temperature (SST), thermocline depth (H), and sub-surface temperature (TSUBA). In the Supplementary Information, we also include the results using the GFDL CM4 model.

\subsubsection{Idealized models}
The idealized model used in this section is a simple intermediate-coupled atmosphere-ocean-SST system~\cite{chen2023simple} (hereafter referred to as the CF23 model), which consists of a set of stochastic partial differential equations. The model is coarse-grained in three key aspects: (1) it describes only a subset of the variables resolved by operational models, (2) it is confined to the equator, reducing it to one spatial dimension, and (3) it has a coarser zonal resolution of approximately $3^o$ ($49$ grid points across $140^\circ$ to $280^\circ$). For developing the bridging model with CESM2, we use only the equatorial SST and thermocline depth from the CF23 model.

In the Discussions Section, we will further present results using an even simpler conceptual model~\cite{chen2022multiscale} (hereafter referred to as the CFY22 model). The CFY22 model comprises just six stochastic ordinary differential equations and represents the SST field using only two variables, averaged over the eastern (Ni\~no 3) and central (Ni\~no 4) Pacific regions. Notably, these are the only two variables used for developing the bridging model with CESM2.

For both the CF23 and CFY22 models, the SST time series and statistics in the Ni\~no 3 and Ni\~no 4 regions closely resemble those from the GODAS reanalysis. These idealized models are capable of generating diverse large-scale spatiotemporal patterns, including extreme El Ni\~no events. The details of both the CF23 and CFY22 models are included in the Supplementary Information.

\subsubsection{Machine learning configuration and training}
The autoencoder's latent space was designed as a vector of size 501, which captures 99\% of the system energy (see the Supplementary Information for details). This was augmented with 98 physical variables (SST and H at 49 equatorial grid points each) provided as pseudo-observations by the CF23 idealized model. The autoencoder training required approximately 3 hours, achieving a final reconstruction loss of $\mathcal{L}_{\text{recon}} \approx 10^{-5}$ and a correlation loss of $\mathcal{L}_{\text{corr}} = -0.6$. The LSTM forecast model required about 4 hours to achieve a comparable reconstruction loss. All computations were performed on CUDA-enabled GPUs.

\subsubsection{Data assimilation setup}
The data assimilation uses 50 ensembles in the EnKF. To ensure stability and physical consistency, we apply multiplicative inflation with a coefficient of $1.09$ to mitigate ensemble collapse and us\\
e the Gaspari-Cohn correlation function for covariance localization to suppress spurious long-range correlations. Pseudo-observations from the idealized models are perturbed according to the specified observation error covariance. The technical details are included in the Supplementary Information.

\subsection{ENSO bias correction in the bridging model}
\subsubsection{Improved representation of ENSO spatial structure}
\begin{figure}[htbp]
    \centering
    \includegraphics[width=1\textwidth]{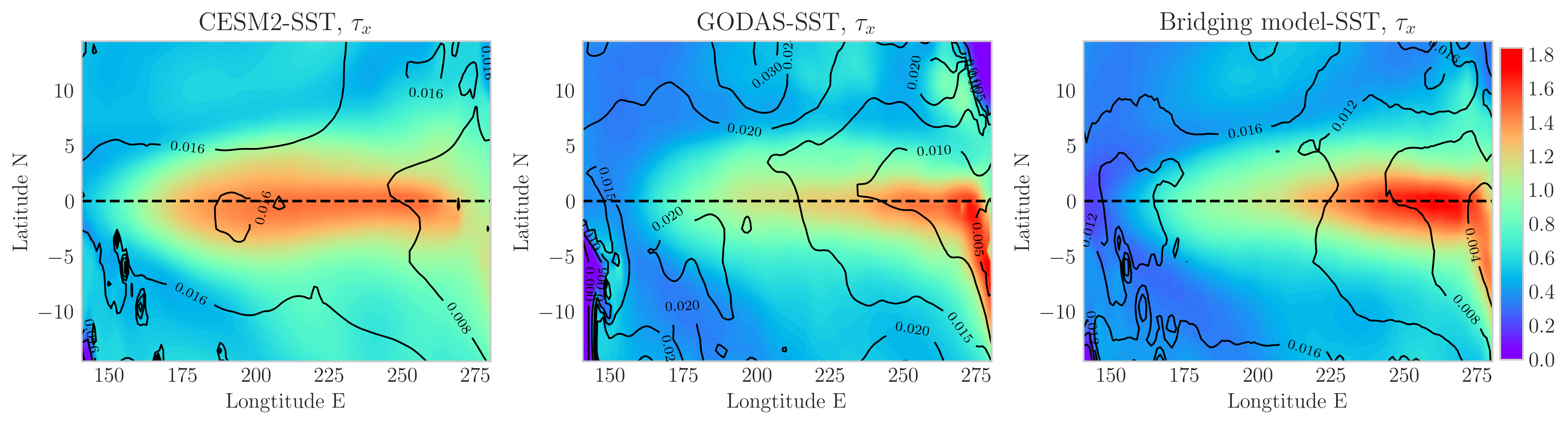}
    \includegraphics[width=1\textwidth]{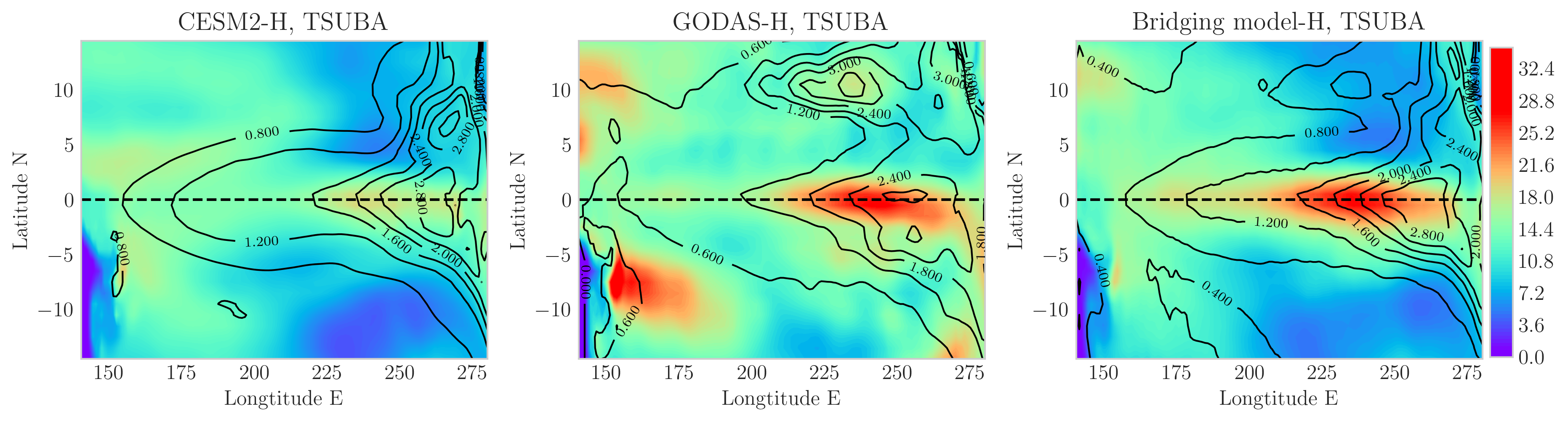}
    \caption{Comparison of different variability patterns, measured by the standard deviation, among CESM2, GODAS reanalysis, and the proposed bridging model. The left, middle, and right columns show results from CESM2, GODAS, and the bridging model, respectively. The top row displays sea surface temperature (SST, shading) and zonal wind stress ($\tau_x$, contours). The bottom row shows thermocline depth ($H$, shading) and subsurface temperature (TSUBA, contours).}
    \label{fig:memory_LSTM}
\end{figure}

Figure \ref{fig:memory_LSTM} compares different variability patterns, measured by the standard deviation, among CESM2, GODAS reanalysis, and the proposed bridging model. In the top row, the spatial distribution of SST variability from GODAS (shading) is strongest in the EP. It exhibits a meridionally asymmetric structure, with a more pronounced signal in the southern hemisphere. In contrast, the CESM2 simulates an SST pattern shifted towards the CP, with similar strength in the Ni\~no 3 and Ni\~no 4 regions. Likewise, the zonal wind stress (contours) in CESM2, particularly in the western Pacific, is generally weaker than in GODAS. These discrepancies suggest that CESM2 struggles to accurately simulate the distinct characteristics of EP and CP El Ni\~no events. Furthermore, CESM2's off-equatorial signals are much weaker than those in the reanalysis. The bridging model significantly improves upon these results. Both the SST and zonal wind stress patterns closely resemble those from GODAS. Similarly, the bottom row reveals a substantial gap between CESM2 and GODAS in simulating thermocline depth and subsurface temperature. The bridging model successfully reproduces the spatial patterns almost perfectly within the core equatorial band (5$^o$S–5$^o$N). While it misses some features farther from the equator, such as the thermocline depth in the southwestern Pacific and subsurface temperature in the northeastern Pacific, the overall structure is markedly improved compared to CESM2. These remaining biases are due to the weak correlation between the equatorial region and these off-equatorial areas, which limits the ability of the data assimilation to effectively propagate information from the idealized model along the equator to the entire domain.

\subsubsection{Propagated improvements in SST and ENSO indices}
\begin{figure}[htbp]
    \centering
    \includegraphics[width=1\textwidth]{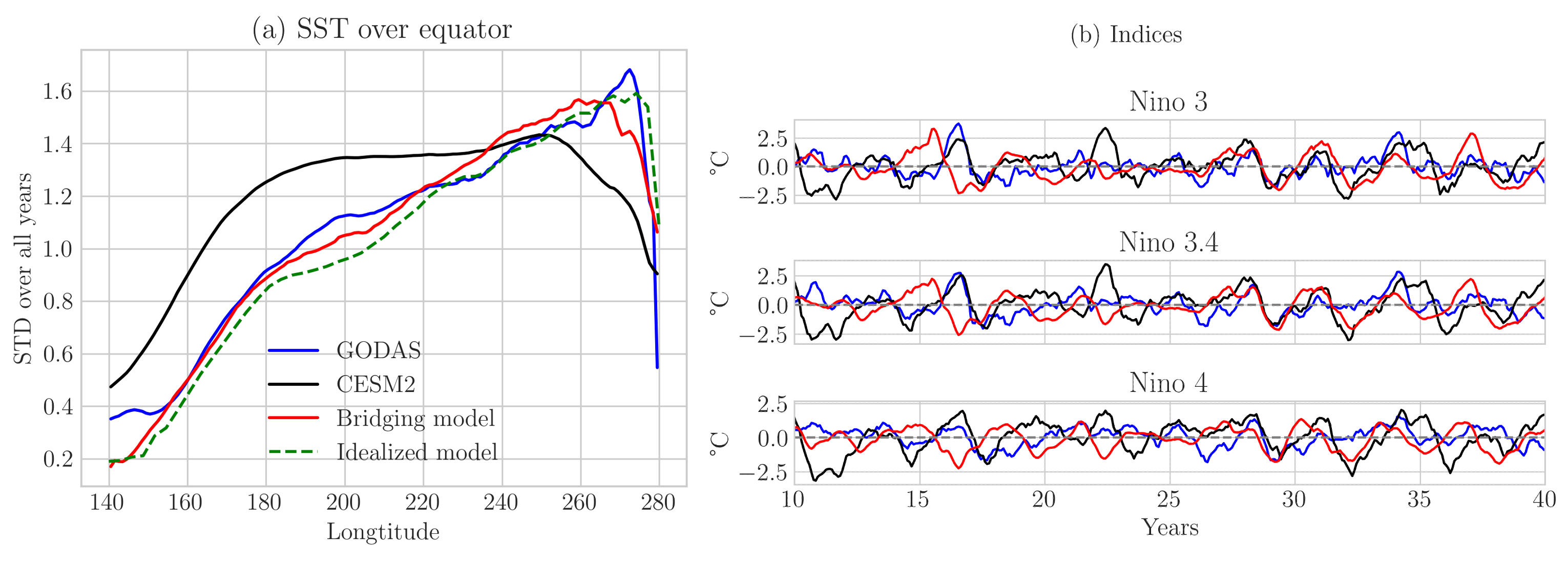}

    \caption{Comparison of equatorial Pacific SST and Ni\~no indices from the CESM2 model, GODAS reanalysis, and the bridging model. The left panel shows the SST along the equator (where the result using the idealized model is also included), while the right panel displays the Ni\~no 3, Ni\~no 3.4, and Ni\~no 4 indices.}
    \label{fig:Assimilated_equator_inflated_sst}
\end{figure}

The idealized model provides SST data only along the equator. However, the commonly used Ni\~no indices are defined as an average over a longitudinal band and a latitudinal range from 5$^o$S to 5$^o$N. Therefore, it is essential to examine both the SST along the equator and the meridionally-averaged SST. Figure \ref{fig:Assimilated_equator_inflated_sst} shows SST along the equator and Ni\~no indices from the CESM2 model, GODAS reanalysis, and the bridging model. The left panel shows that the equatorial SST of the bridging model closely matches that of the GODAS reanalysis, with the signal intensifying towards the EP. In contrast, the SST from CESM2 is generally stronger than GODAS and remains uniformly high from the CP to the EP. Similar results are obtained for the SST averaged from 5$^o$S to 5$^o$N, which are displayed in the Supplementary Information. The right panel in Figure \ref{fig:Assimilated_equator_inflated_sst} shows the corresponding time series for the three Ni\~no regions. The bridging model produces a qualitatively similar profile to GODAS, with more pronounced positive phases (El Ni\~no events) than negative phases (La Ni\~na events), especially in the Ni\~no 3 index. Conversely, CESM2 exhibits a more symmetric oscillation in SST amplitude. Its time series is also smoother and displays more regular events with a roughly three-year period.

\subsubsection{Correcting ENSO statistics}
\begin{figure}[htbp]
    \centering
    \includegraphics[width=1\textwidth]{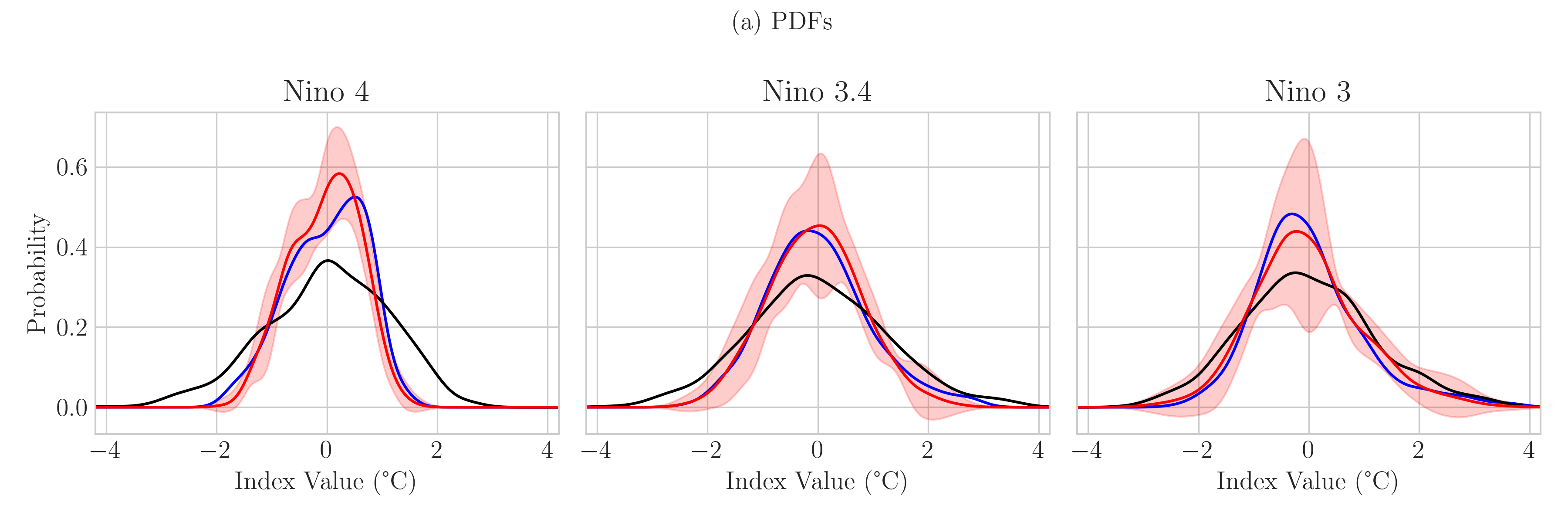}
    \includegraphics[width=1\textwidth]{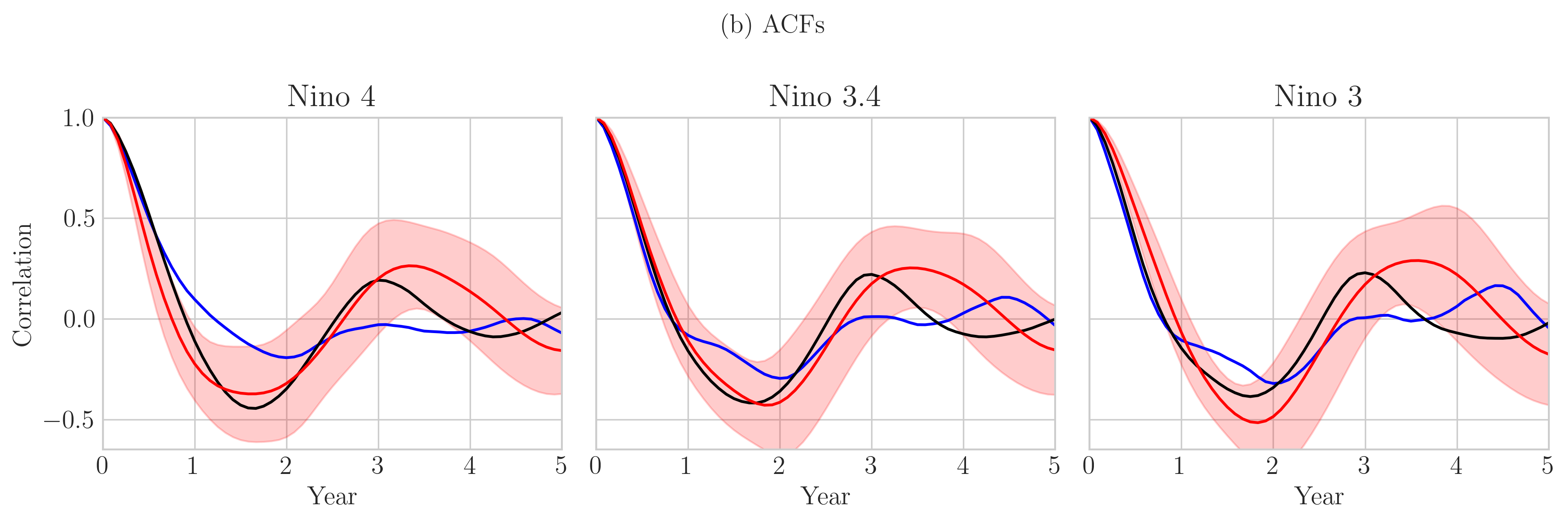}
    \includegraphics[width=1\textwidth]{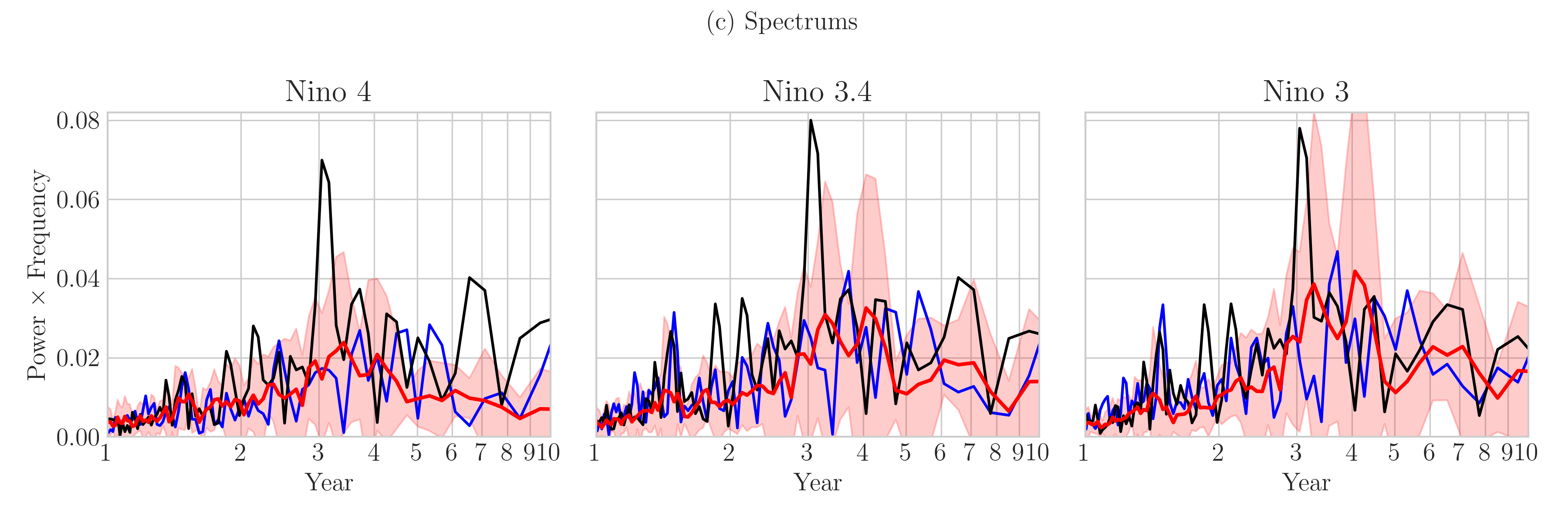}
    \includegraphics[width=1\textwidth]{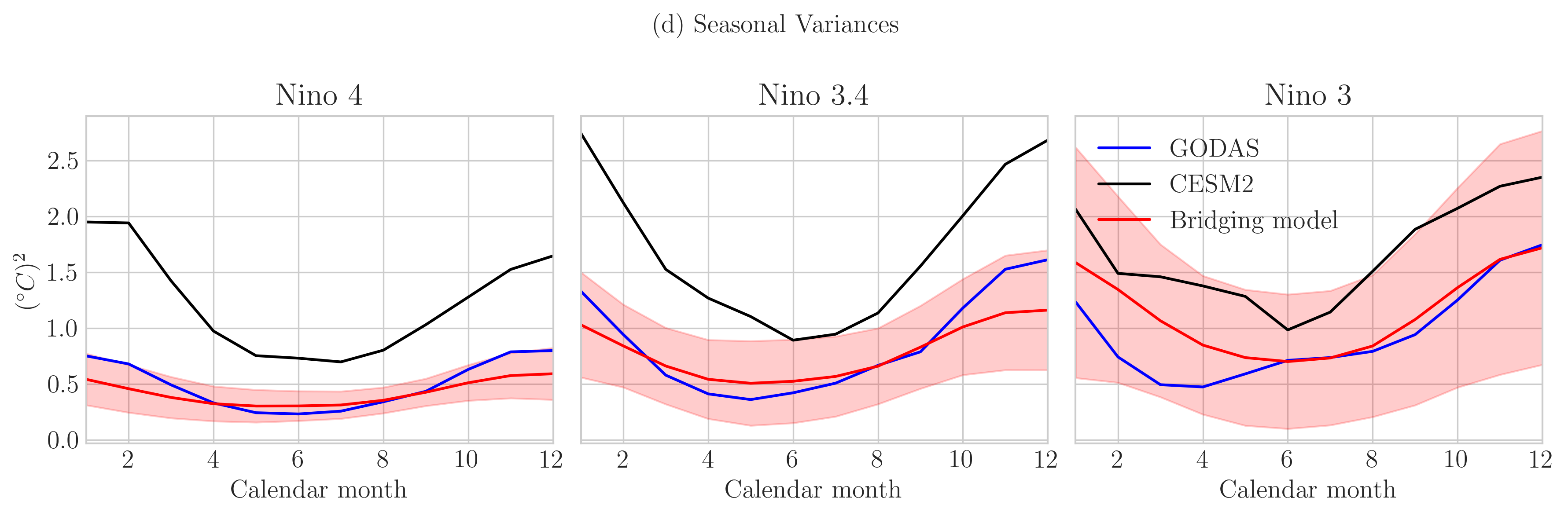}
    \caption{Statistical properties of the Ni\~no 3, Ni\~no 3.4, and Ni\~no 4 indices for the CESM2 model, GODAS reanalysis, and the bridging model. The four rows present the probability density functions (PDFs), autocorrelation functions (ACFs), power spectra, and seasonal variation of SST. For the bridging model, the red shaded area represents the 95\% confidence interval derived from 100 ensemble members, each with a length equal to the 40-year GODAS reanalysis period. }
    \label{fig:stats}
\end{figure}

Figure \ref{fig:stats} compares the statistical properties of the Ni\~no 3, Ni\~no 3.4, and Ni\~no 4 indices across the different models. Consistent with the previous figures, the results of the bridging model largely resemble those of the GODAS reanalysis. The bridging model produces nearly identical probability density functions (PDFs) to those of the reanalysis data. It captures the positive and negative skewness in the Ni\~no 3 and Ni\~no 4 indices, respectively, which correspond to the asymmetry in both EP and CP ENSO events. The autocorrelation functions (ACFs) for the Ni\~no 3.4 and Ni\~no 3 regions are also reproduced with high accuracy. Notably, the ACF of the bridging model decays slightly faster than that of GODAS, which is a bias inherited from the idealized model. Next, both the GODAS reanalysis and the bridging model exhibit a broad power spectrum, indicative of the intrinsic turbulent nature of this interannual variability. In contrast, CESM2 shows a strong spectral peak at approximately 3 years, which is consistent with its overly regular time series evident in Figure \ref{fig:Assimilated_equator_inflated_sst}. Finally, the bridging model accurately captures the SST seasonal variability, indicating more active ENSO events happening in boreal winter, whereas CESM2 consistently overestimates both the seasonal and total variance.

\subsubsection{Realistic simulation of ENSO diversity and complexity}
\begin{figure}[htbp]
    \centering
    \includegraphics[width=1\textwidth]{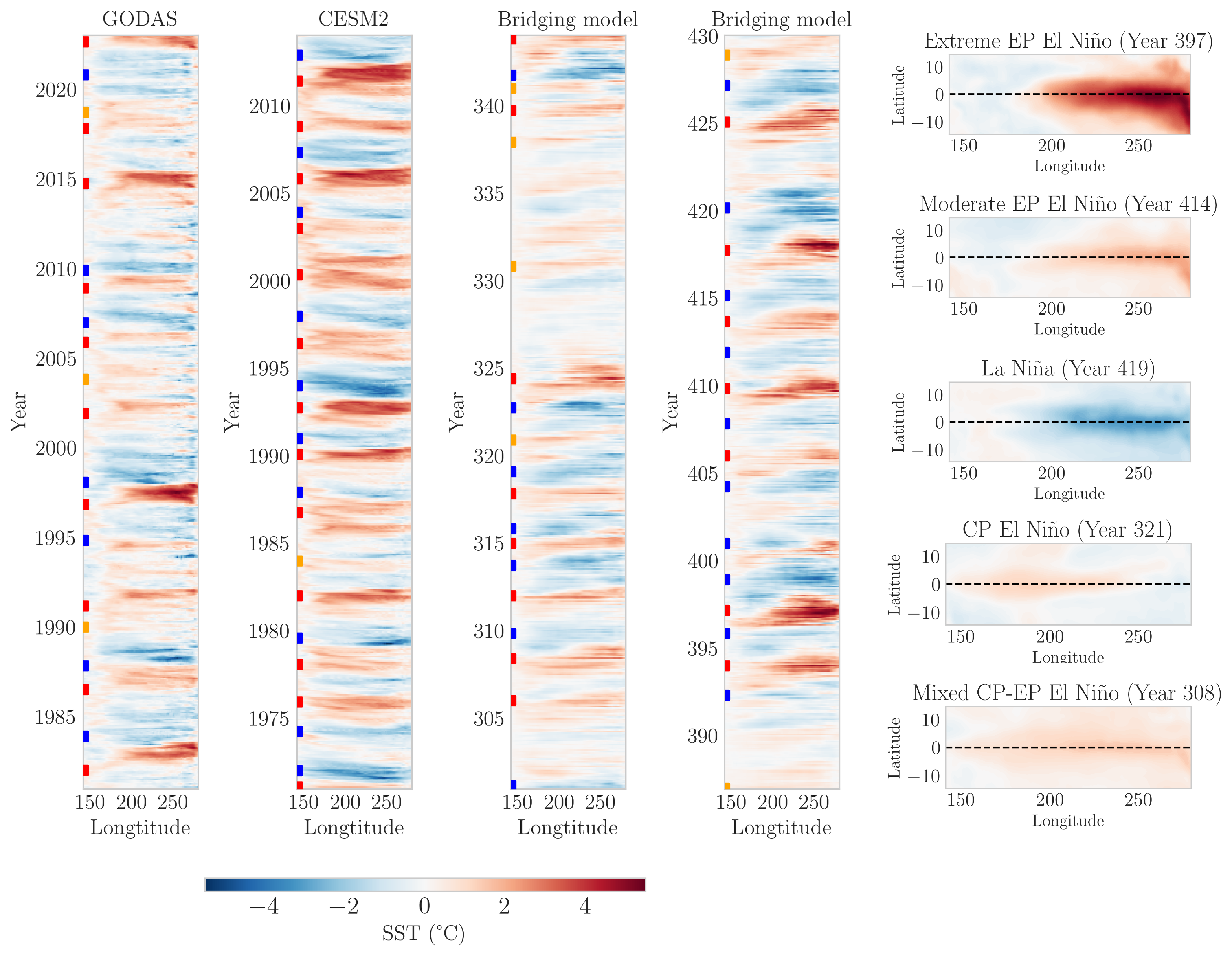}

    \caption{Comparison of SST evolution along the equator: Hovm\"oller diagrams for CESM2, GODAS, and the bridging model (left) and snapshots of distinct ENSO events (right). The red squares indicate the EP El Ni\~no, the blue ones show the LA Ni\~no, and the orange squares represent the CP El Ni\~no events. }
    \label{fig:Var_HOV_SST}
\end{figure}

Figure \ref{fig:Var_HOV_SST} compares the time evolution of equatorial SST anomalies in the CESM2 model, GODAS reanalysis, and the bridging model via the Hovm\"oller diagrams. As expected, CESM2 has difficulties distinguishing between different types of El Ni\~no events, with most events occurring across both the CP and EP regions. In contrast, the bridging model generates a more diverse range of events, including extreme EP El Ni\~no, moderate EP El Ni\~no, CP El Ni\~no, mixed EP-CP El Ni\~no, and La Ni\~na events. It also successfully produces multi-year El Ni\~no and La Ni\~na events, as in the GODAS reanalysis. These results demonstrate that the bridging model skillfully reproduces not only the observed key statistical features but also the realistic simulations of individual events. The Hovm\"oller diagrams of other variables can be found in Supplementary Information.


\section{Discussions}
\subsection{Computational advantages} The XAI framework developed here demonstrates that the skillful integration of a model hierarchy can significantly improve simulation accuracy. A key practical advantage of this method is its computational efficiency. Operational models are essential for sensitivity experiments and future scenario projections. However, their high computational cost poses a significant constraint. The bridging model overcomes this limitation by employing dimension reduction in latent data assimilation and utilizing a machine learning surrogate as the short-term forecast model in data assimilation, thereby advancing rapid execution. The bridging model completes a 42-year simulation (matching the GODAS record) in just 8 minutes and 20 seconds using PyTorch and running on Google Colab's CPU environment, which offers 12 GB of RAM and two vCPUs (virtualized Intel Xeon at 2.3 GHz). This efficiency, combined with the advantage of utilizing the idealized model in generating long simulations, allows the bridging model to produce large ensembles. Such extensive sampling is crucial in Earth science, where the observational record is short, as it significantly improves our understanding of statistical behavior, underlying mechanisms, and uncertainties, particularly for extreme events.

\subsection{Sensitivity tests and digital twins} From a broader perspective, the bridging model can function as a digital twin for sensitivity analysis and forecasting with inherent uncertainty quantification, answering the ``what-if'' questions. This can be achieved in several ways. One straightforward way is to train a set of bridging models based on the operational model under different external forcings. A digital twin can then be constructed with the forcing value as an input, enabling scenario prediction for interpolated forcings. Nevertheless, a more efficient method, which is uniquely suited to this framework, is to run sensitivity experiments within the cheap idealized model (e.g., modifying decadal variability that modulates ENSO event types) and then assimilate its output into the operational model's surrogate via the latent space. This method requires training the surrogate model only once. The response of different physical fields can then be analyzed using the various outputs of the bridging model. Such a method, with the assistance of the cheap idealized model, overcomes the fundamental difficulties of extrapolation in purely machine learning based digital twins. Notably, because the bridging model state is estimated through data assimilation, it naturally contains the uncertainty.

\begin{figure}[htbp]
    \centering
    \includegraphics[width=.8\textwidth]{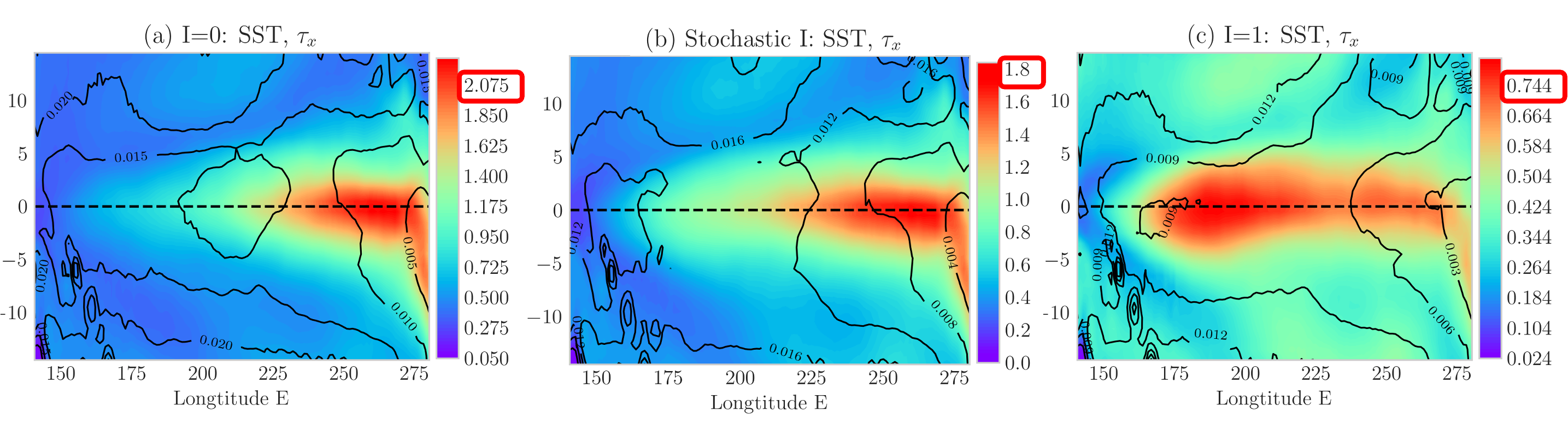}
\includegraphics[width=.8\textwidth]{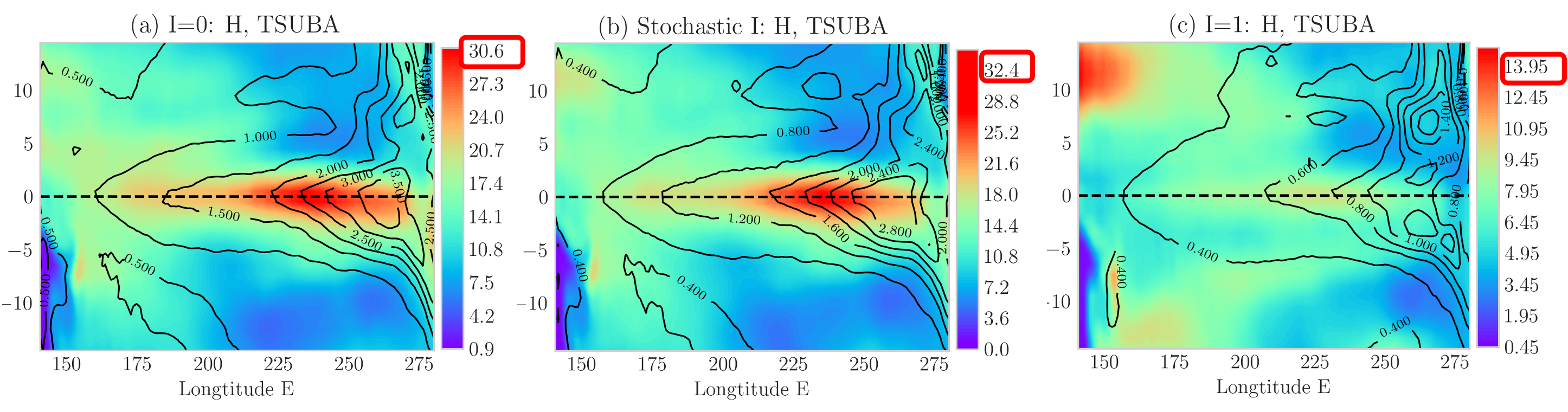}
    \caption{Sensitivity of the bridging model's variability to the decadal variable in the idealized model (CF23), $I$, a surrogate for Walker circulation strength. In the CF23 model, $I$ is a stochastic process bounded by $0$ (weakened Walker circulation) and $1$ (strengthened circulation). The standard deviation of key fields is shown for three regimes: a persistently weak ($I=0$, left), a stochastic ($0\leq I \leq1$, middle), and a persistently strong ($I=1$, right) Walker circulation. The top row displays sea surface temperature (SST, shading) and zonal wind stress ($\tau_x$, contours); the bottom row shows thermocline depth (H, shading) and subsurface temperature (TSUBA, contours). For clarity of presentation, the colorbar maximum is adjusted independently for each regime.}
    \label{fig:Sens_SST}
\end{figure}

Figure \ref{fig:Sens_SST} presents ``what-if'' scenarios from the bridging model, simulating a persistently weakened (left column) and strengthened (right column) Walker circulation. This circulation is controlled by the decadal variable $I$ in the CF23 idealized model. A weakened Walker circulation ($I=0$) leads to a higher prevalence of EP and extreme El Ni\~no events. Consequently, the bridging model simulates an eastward shift in SST variability, and an increase in amplitude. This change is accompanied by a slight intensification of intraseasonal wind stress, subsurface temperature, and thermocline depth in the EP. Conversely, a strengthened Walker circulation ($I=1$) favors CP events, resulting in a westward shift of the SST pattern and a general weakening of the other fields. These results are consistent with previous observational and model studies~\cite{mcgregor2014recent, yu2012identifying}, but the bridging model achieves these what-if scenarios with significantly greater computational efficiency.

\subsection{Robustness of the bridging model} Figure \ref{fig:concept_SST} presents the bridging model results using only two SST time series from a 6-dimensional stochastic ODE (CFY22) as the idealized model. Although this is an oversimplified minimal-order representation, the bridging model, using the same surrogate forecast model trained on CESM2, successfully overcomes the model barrier. This demonstrates that even an extremely simple yet statistically accurate idealized model can be leveraged by the XAI framework to generate significant improvements. Furthermore, it confirms the robustness of the results with respect to the choice of the underlying idealized model.

\begin{figure}[htbp]
    \centering
    \includegraphics[width=.8\textwidth]{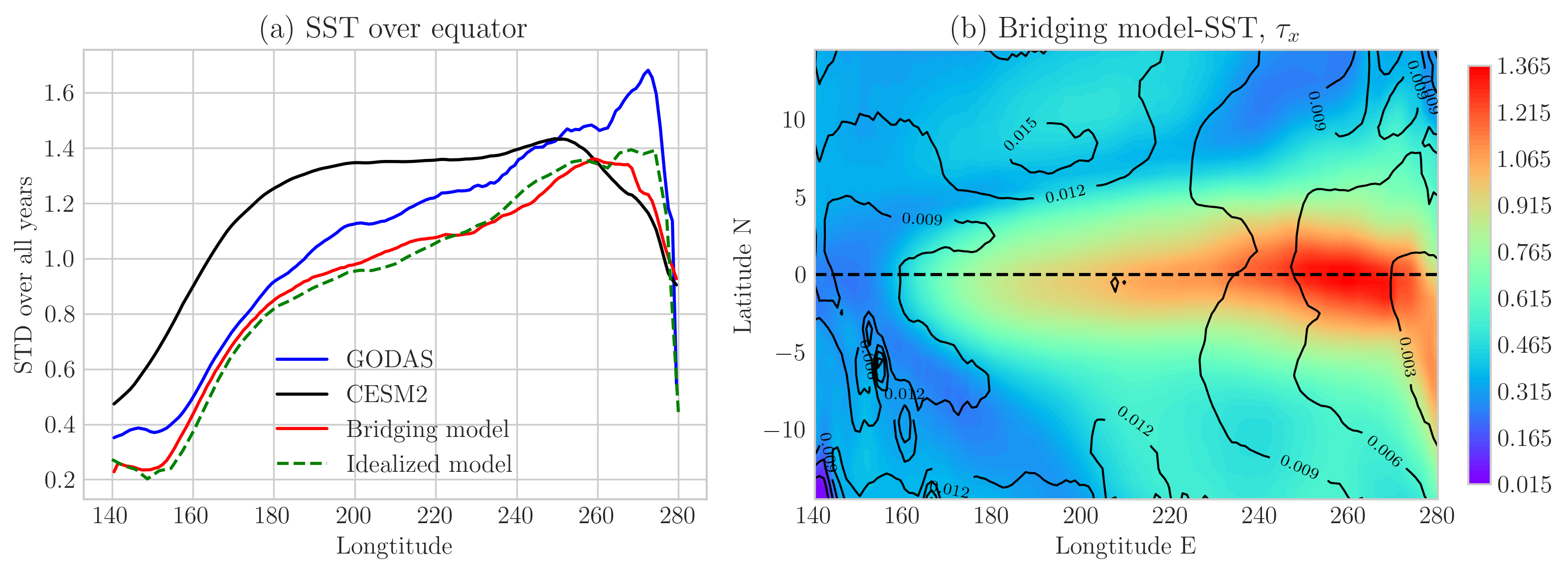}

    \caption{Comparison of equatorial Pacific SST, measured by the standard deviation,
among CESM2, GODAS reanalysis, and the bridging model with pseudo-observations provided by even a simpler 6-dimensional stochastic ODE model (CFY22 model; left panel). The right panel
displays (SST, shading) and zonal wind stress ($\tau_x$, contours).  }
    \label{fig:concept_SST}
\end{figure}

The results in the main text are all based on CESM2. In the Supplementary Information, the results of the bridging model based on the GFDL CM4 using both the CF23 and CFY22 idealized models are also included. The bridging model consistently improves the performance compared to the operational models, further demonstrating the robustness of the framework.

\subsection{Multi-model scenarios}
Although the present study uses a single operational and idealized model, the framework can be readily generalized to multi-model scenarios. A promising extension would involve a preprocessing step to identify the strengths of different models in characterizing distinct event types (e.g., EP vs. CP El Ni\~no) or regional features. Incorporating such a diverse model hierarchy is expected to further improve the accuracy of the bridging model. In fact, the residual errors in the current bridging model, primarily its difficulty capturing off-equatorial behavior, can be attributed to a corresponding shortcoming in CESM2. This specific limitation is anticipated to be overcome by integrating operational models that better represent off-equatorial processes.

\subsection{Communication between modeling communities} One of the central components of the XAI framework developed here is the incorporation of an appropriate idealized model. This work, therefore, highlights the critical importance of advancing deeper communication and collaboration between the communities developing high-resolution operational models and those focused on idealized or conceptual frameworks. Traditionally, these groups have operated in relative isolation, with idealized models primarily used for hypothesis testing and conceptual understanding, rather than as direct inputs for improving operational forecasts. Our framework provides a concrete and actionable way through which the targeted accuracy of a well-calibrated idealized model can be systematically translated into stable corrections for a complex operational model. This creates a powerful incentive for idealized model developers to refine their models to capture key statistical features and dynamical processes. Conversely, the method developed here encourages operational modelers to identify specific biases in their systems that could be targeted by simpler models. By bridging these disciplinary gaps, the framework facilitates a synergistic feedback loop, thereby accelerating progress in Earth system modeling as a whole.

\subsection{General use of the output from the bridging model} The XAI framework developed here is not limited to Earth science. It presents a generalizable strategy for integrating multi-fidelity models in any complex dynamical system, where models of varying complexity provide complementary insights. The output generated by the bridging model is not merely a corrected version of the operational simulation. It is a new hybrid data product that inherits the high-resolution and comprehensive physical fields of the operational model while being globally informed by the targeted statistical and dynamical accuracy of the idealized model. This makes the output uniquely valuable for a wide range of applications. The massive ensembles that can be efficiently generated are crucial for robust uncertainty quantification, particularly for assessing the statistics of extreme events that are poorly sampled in the short observational record. Furthermore, the physically consistent high-resolution fields enable detailed process studies that were previously impeded by model biases. For instance, the atmospheric teleconnections and remote impacts of different types of ENSO events can be investigated with greater confidence. The bridging model effectively serves as a powerful tool for generating reliable ``what-if'' datasets, supporting applications ranging from seasonal forecasting and climate projection to educating the next generation of scientists on the realistic dynamics of the Earth system.

\section*{Data availability}
 Data for sea surface temperature, thermocline depth, zonal wind stress, and subsurface temperature at 50 meters from both observations and operational models are used in this study. All variables can be directly downloaded from their respective repositories with the exception of the thermocline depth, which is calculated as the depth of the 20$^{\circ}$C isotherm. Anomaly data is used here which is found by subtracting the monthly climatological mean. Observational data are sourced from the GODAS reanalysis data set, which covers the time period from 1981 to 2023 \cite{behringer2004evaluation}. The operational model data from CESM2 and GFDL-CM4 is historical, spanning from 1850 to 2014 \cite{eyring2016overview}. The operational model data is constricted to the equatorial Pacific region over latitudes 15$^\circ$S-15$^\circ$N and longitudes 140$^\circ$E-80$^\circ$W to match the GODAS data set. All relevant processed data in this study and the code can be found in~\href{https://github.com/pbehnoud/bridging_model}{GitHub repository}.

\section*{Acknowledgment}
The research of N.C. is funded by the Office of Naval Research N00014-24-1-2244 and the Army Research Office W911NF-23-1-0118. P.B. and C.M. are partially supported as a research associate and a research assistant under
the second grant. S.C. acknowledges the support of the French Agence Nationale de la Recherche (ANR) under reference ANR-22-CPJ2-0143-01. M.B. and S.C. acknowledge the support of France 2030 PEPR Maths-Vives, through the DRUIDS project, grant ANR-24-EXMA-0002. CEREA is a member of Institut Pierre-Simon Laplace (IPSL).

\section*{Author contributions}

N.C. designed the project. P.B. developed and implemented the algorithms. N.C. and P.B. wrote the manuscript. M.C. processed the data, contributed to the physical interpretation of numerical studies, and helped design experiments. S.C. assisted with the implementations. M.B. contributed to revisions and discussions. All authors contributed to discussions, reviewed, and edited the manuscript.

\section*{Competing interests}
The authors declare no competing interests.

\section*{Additional information}

The Supplementary Information is available. 
\bibliography{references2}

\newpage

\clearpage
\appendix

\renewcommand{\thesection}{S.\arabic{section}}
\renewcommand{\thefigure}{S.\arabic{figure}}
\renewcommand{\thetable}{S.\arabic{table}}
\renewcommand{\theequation}{S.\arabic{equation}}

\setcounter{section}{0}
\setcounter{figure}{0}
\setcounter{table}{0}
\setcounter{equation}{0}

	\begin{center}
		{\huge Supplementary Information\\
			Bridging Idealized and Operational Models: An Explainable AI Framework for Earth System Emulators}\medskip
			
		\vspace{.3cm}
		
		{\large Pouria Behnoudfar$^1$, Charlotte Moser$^1$, Marc Bocquet$^2$, Sibo Cheng$^2$, Nan Chen$^{1*}$}\\
		{\large $^1$Department of Mathematics, University of Wisconsin-Madison, Wisconsin, USA.}\\
		{\large $^2$CEREA, ENPC, EDF R\&D, Institut Polytechnique de Paris,
			Île-de-France, France.}\\
		{\large $^*$Corresponding author's email: chennan@math.wisc.edu}
	\end{center}
			\vspace{.5cm}
	\section{Data and network structures for ENSO application}
	
	Before feeding the data into the LSTM encoder-decoder network, each input variable is normalized with respect to 1.2 times its maximum value observed across the time interval. This normalization ensures that all inputs are scaled within a range suitable for the \texttt{tanh} activation used in the last layer of both the encoder, LSTM, and decoder, improving training stability. Additionally, it provides flexibility for downstream data assimilation, where the normalized range can be temporarily expanded by up to 20\% if a stronger signal representation is required. Furthermore, by controlling the magnitude of the inputs, the network prevents excessive saturation of activation functions and preserves gradient flow, facilitating more accurate latent representations and sequence predictions.
	
	Next, to develop a reduced-space surrogate model, we use an encoder that compresses the input into a low-dimensional latent representation. Then, we enrich the latent space using physical variables defined on the observed grid points, while the decoder reconstructs the augmented space back to the original spatial resolution.
	
	\subsection{Autoencoder structure}
	The architecture for the ENSO application is summarized below.
	
\begin{figure}[H]
		\centering
		\includegraphics[width=.6\linewidth]{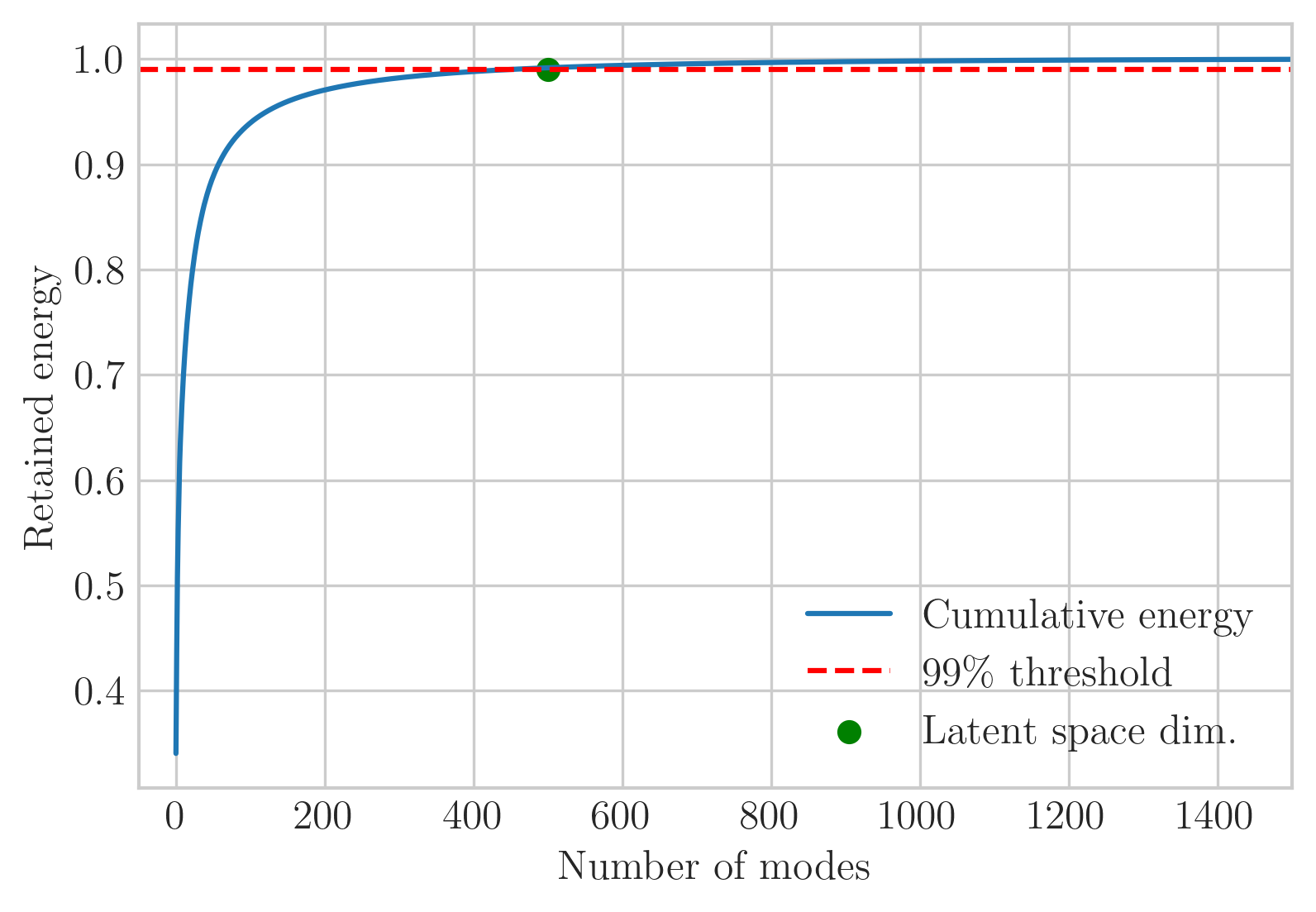}
		
		\caption{Determining the latent space dimension using the PPO.  }
		\label{fig:POD}
	\end{figure}

	The encoder takes an input with $4$ variables and a spatial resolution of $(140,30)$.  It consists of successive convolutional layers with batch normalization, GELU activation, and pooling layers to gradually reduce spatial resolution while increasing the number of feature channels:
	
	\begin{itemize}
		\item Conv2d: $(4,140,30) \to (32,140,30)$, BatchNorm, GELU, MaxPool2d $(2) \to (32,70,15)$
		\item Conv2d: $(32,70,15) \to (64,70,15)$, BatchNorm, GELU, MaxPool2d $(2) \to (64,35,7)$
		\item Conv2d: $(64,35,7) \to (128,35,7)$, BatchNorm, GELU, AdaptiveAvgPool2d $(16,4) \to (128,16,4)$
		\item Fully connected layer: $\mathbb{R}^{128 \times 16 \times 4} \to \mathbb{R}^{n_l}$, followed by $\tanh$ activation. 
	\end{itemize}
	Thus, the encoder maps the input into a latent vector $\mathbf{z} \in \mathbb{R}^{n_l}$ (here, $n_l=501$). Proper Orthogonal Decomposition (POD) is used to determine the size of the latent space based on the number of modes required to capture more than 99\% of the model's energy. We show the cumulative energy with respect to the modes in~\ref{fig:POD}.
	
	The decoder reconstructs the original input shape from the latent vector and the augmented physical variables.
	It first maps $\mathbf{x}=[\mathbf{z} ,\, \mathbf{y}]^T$ back into a tensor of shape $(128,16,4)$ using a fully connected layer, and then applies a sequence of transposed convolutions to increase spatial resolution while reducing channel dimension:
	
	\begin{itemize}
		\item Fully connected: $\mathbb{R}^{n_l+n_o} \to \mathbb{R}^{128 \times 16 \times 4}$, reshape to $(128,16,4)$
		\item ConvTranspose2d: $(128,16,4) \to (64,32,8)$, BatchNorm, GELU
		\item ConvTranspose2d: $(64,32,8) \to (32,64,16)$, BatchNorm, GELU
		\item ConvTranspose2d: $(32,64,16) \to (16,128,32)$, BatchNorm, GELU
		\item Conv2d: $(16,128,32) \to (4,128,32)$, $\tanh$ activation
		\item Bilinear interpolation to final output resolution $(4,140,30)$
	\end{itemize}
	
\begin{figure}[H]
		\centering
		\begin{subfigure}[b]{\linewidth}
			\includegraphics[width=.9\linewidth]
			{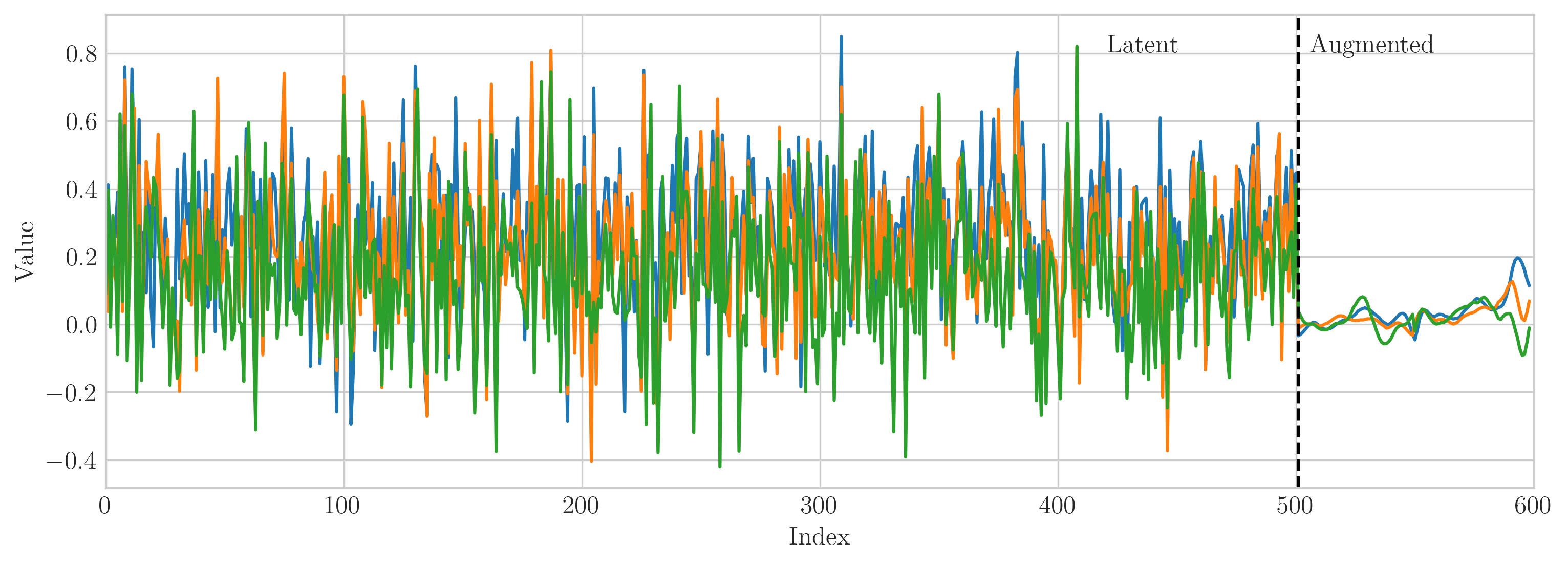}
			\caption{Our approach: choice of $\lambda$ ensures the same order of reconstruction and correlation losses.}
		\end{subfigure}
		\begin{subfigure}[b]{\linewidth}
			\includegraphics[width=.9\linewidth]
			{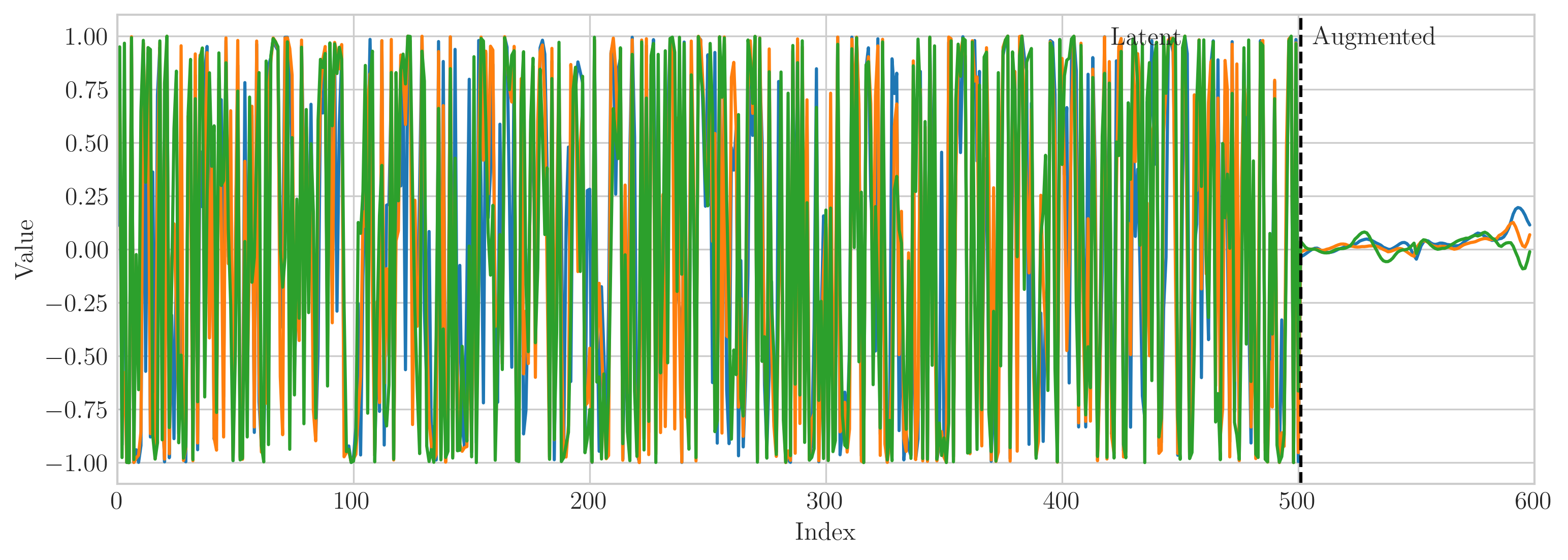}
			\caption{$\lambda=0$, training only using the reconstruction loss.}
		\end{subfigure}
		\begin{subfigure}[b]{\linewidth}
			\includegraphics[width=.9\linewidth]
			{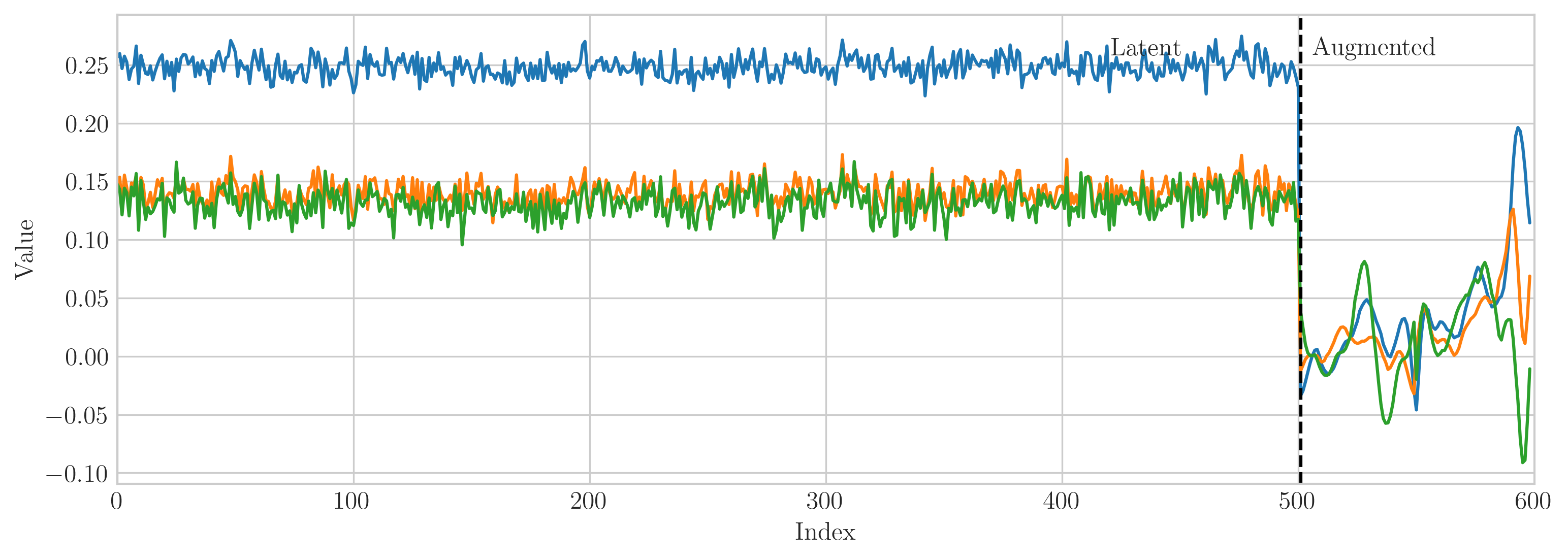}
			\caption{$\lambda=1$, the correlation loss dominates the training process.}
		\end{subfigure}
		\caption{Encoder outputs and corresponding observed physical variables at different time instances.
			In our approach (panel (a)), $\lambda$ is adaptively chosen to maintain the same order in reconstruction and correlation losses. In contrast, when the correlation loss is excluded ($\lambda=0$; panel (b)), the latent space becomes uncorrelated with the observed physical variables, indicating a loss of meaningful correspondence between the learned features and the physical domain. In panel (c), the reconstructed field is far from the ground truth.}
		\label{fig:aug}
	\end{figure}

	Overall, this encoder-decoder design compresses high-dimensional data into a compact latent space and reconstructs it with minimal loss of information, guided by the physical properties of the augmented part. Figure~\ref{fig:aug} shows the encoder's outputs and the physical variables with different values of the $\lambda$ (the weight of the second loss: maximizing the cross-correlation between the latent space and the observables). It is seen that an appropriate choice of $\lambda$ ensures the same order of reconstruction and correlation losses. Otherwise, a large $\lambda$ only focuses on maximizing the correlation between the latent space and the observations while ignoring the accuracy of the latent representation. Similarly, if $\lambda$ is small, the observations may become uncorrelated to the latent variables and thus the data assimilation skill decreases significantly.

\begin{figure}[H]
		\centering
		\begin{subfigure}[b]{\linewidth}
			\includegraphics[width=\linewidth]{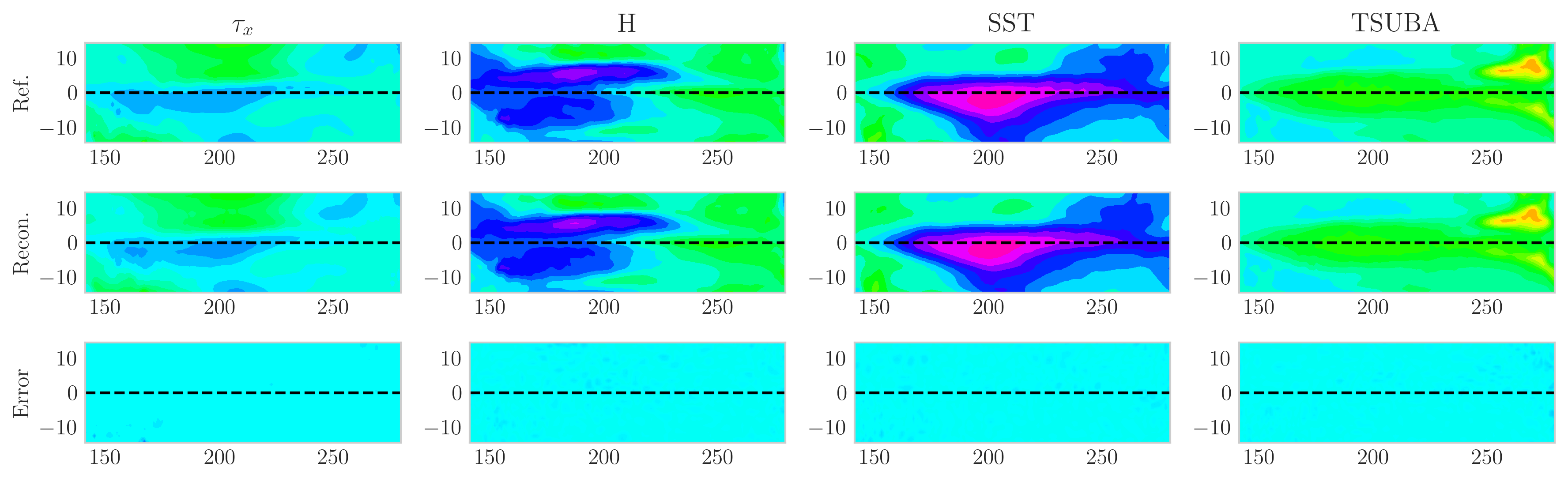}
			\caption{Year 1855}
		\end{subfigure}
		\hfill
		\begin{subfigure}[b]{\linewidth}
			\includegraphics[width=\linewidth]{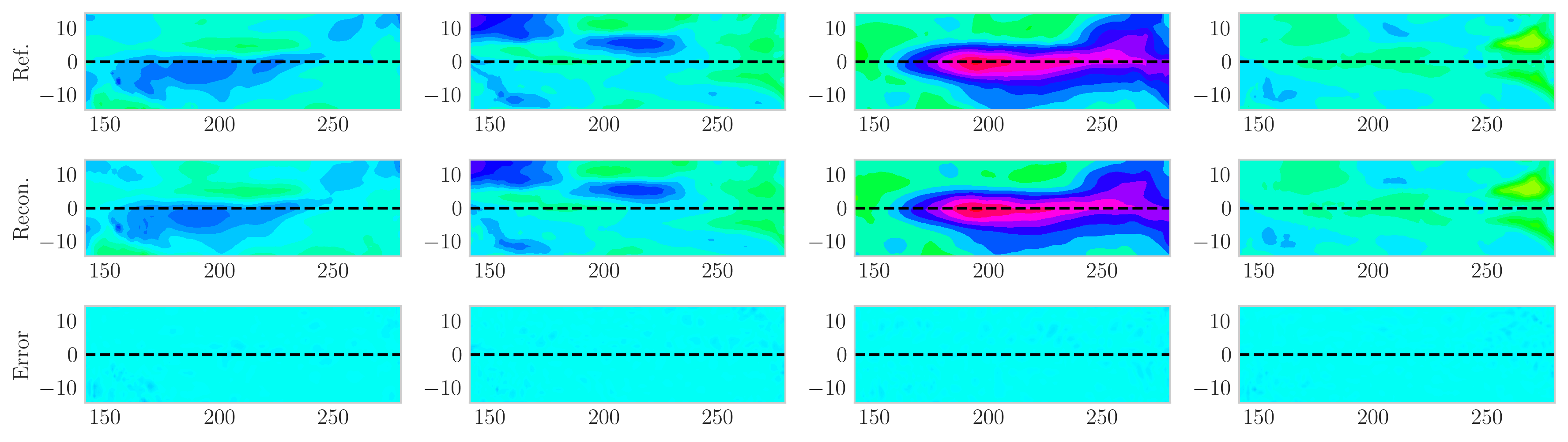}
			\caption{Year 1920}
		\end{subfigure}
		\hfill
		\begin{subfigure}[b]{\linewidth}
			\includegraphics[width=\linewidth]{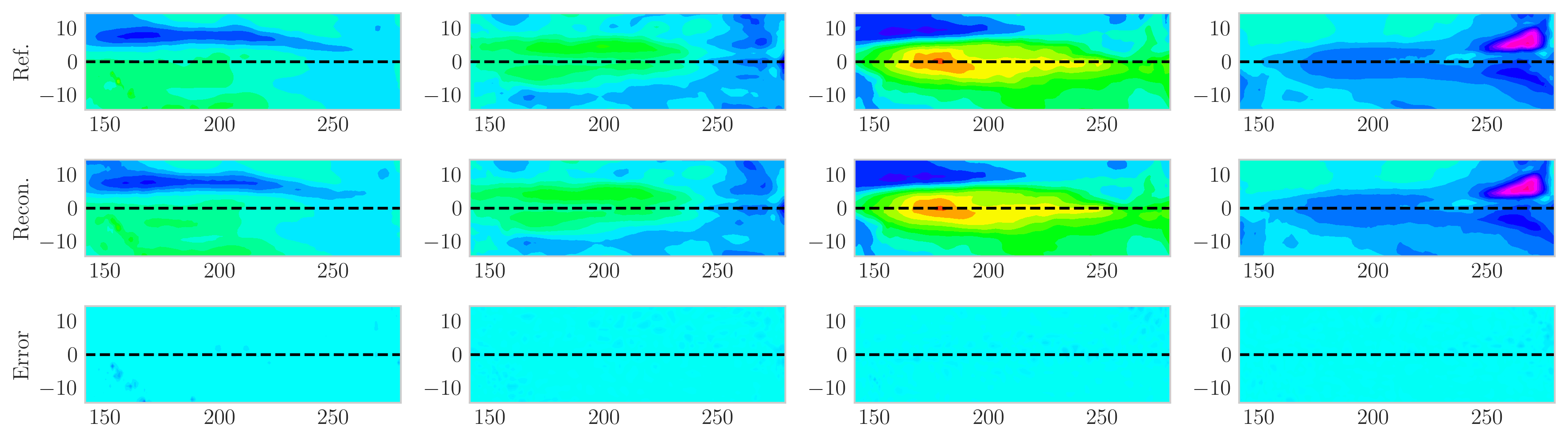}
			\caption{Year 2010}
		\end{subfigure}
		\caption{The accuracy of the reconstructed fields using the autoencoder. }
		\label{fig:recons}
	\end{figure}

	Figure~\ref{fig:recons} compares the reference and the reconstructed field at three time instances with the appropriate choice of $\lambda$, indicating the high consistency between the two fields.

	\subsection{LSTM network structure}
\begin{figure}[H]
		\centering
		\begin{subfigure}[b]{\linewidth}
			\includegraphics[width=\linewidth]{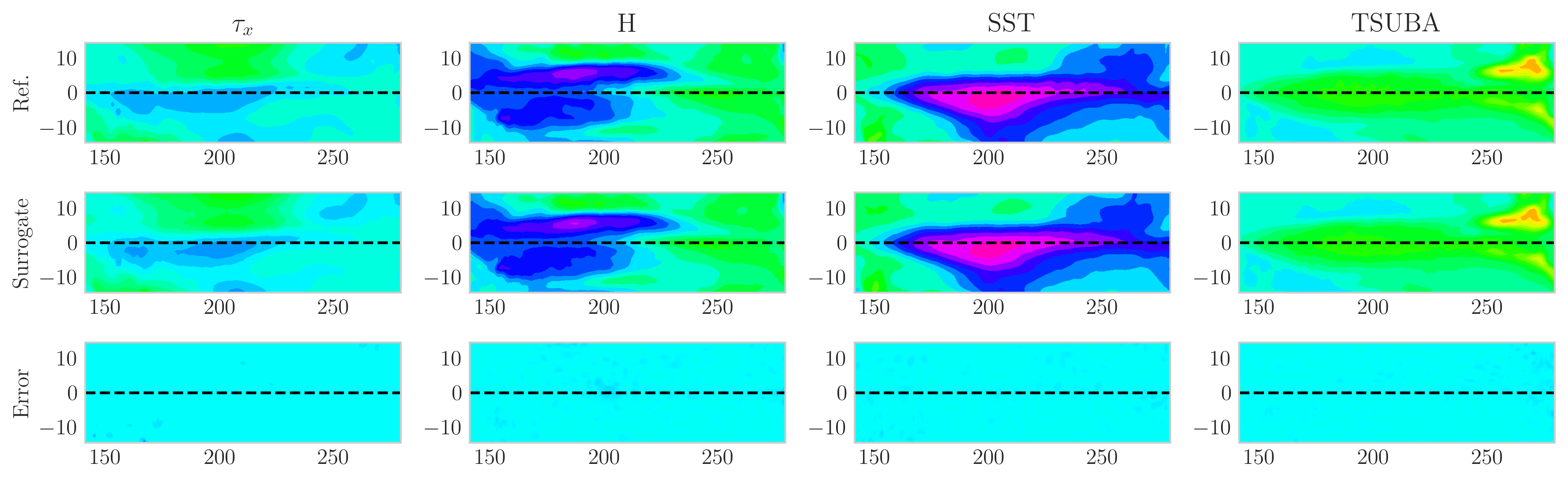}
			\caption{Year 1855}
		\end{subfigure}
		\hfill
		\begin{subfigure}[b]{\linewidth}
			\includegraphics[width=\linewidth]{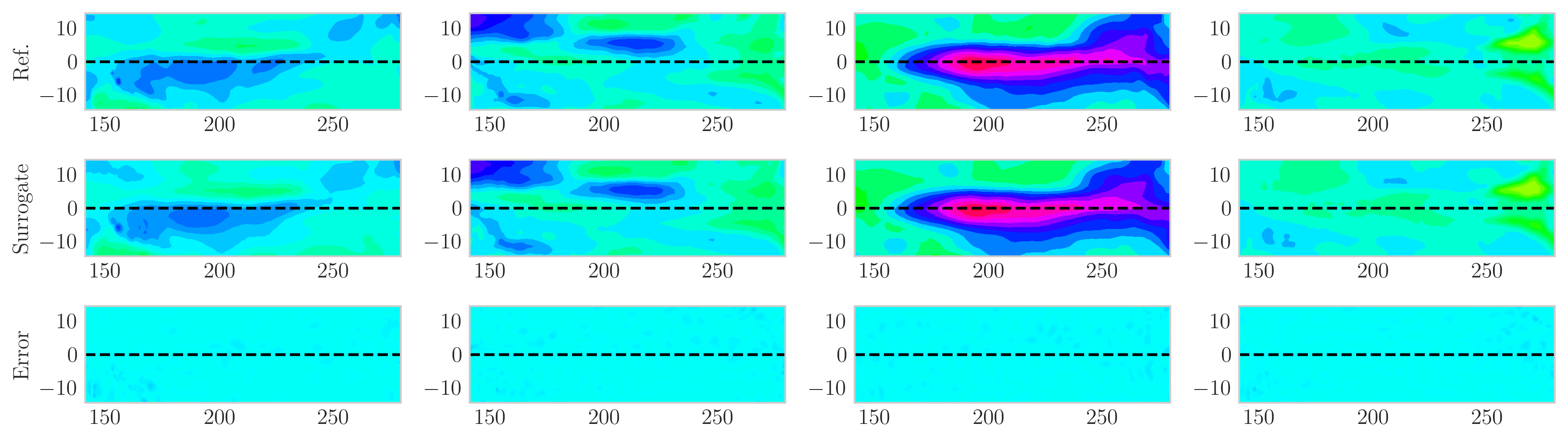}
			\caption{Year 1920}
		\end{subfigure}
		\hfill
		\begin{subfigure}[b]{\linewidth}
			\includegraphics[width=\linewidth]{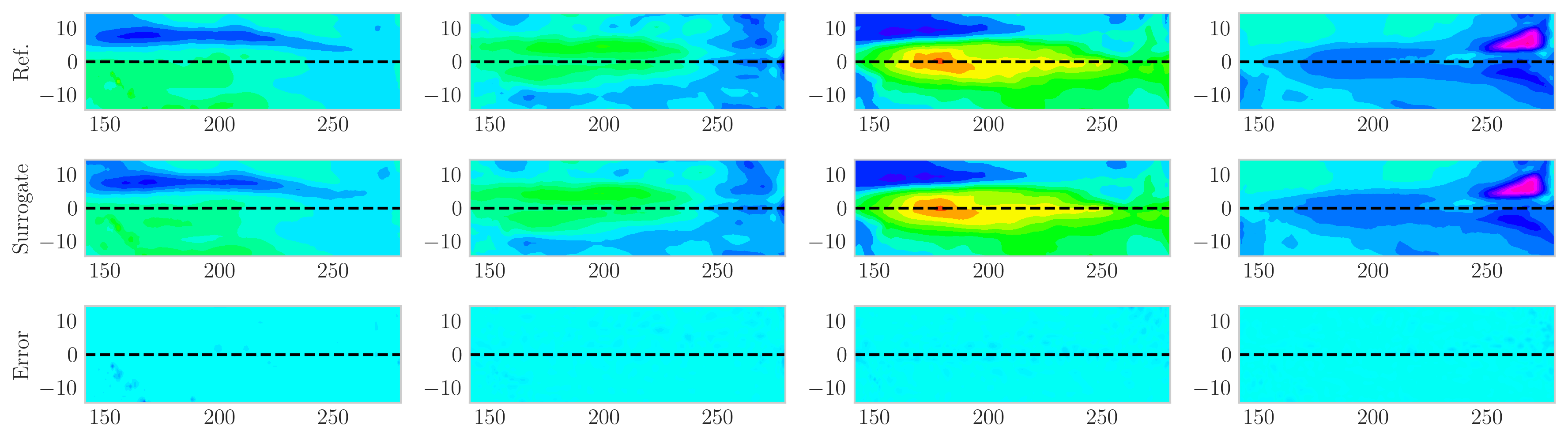}
			\caption{Year 2010}
		\end{subfigure}
		\caption{The accuracy of the predicted fields using the LSTM. }
		\label{fig:lstm}
	\end{figure}

	We employ an LSTM-based network to model the temporal evolution of the latent representation obtained from the encoder.
	The LSTM predicts sequences of length one given an input sequence of latent and observation features within the last two time steps (months). We compare the ground truth and the predicted field at a single time step (one month) in Figure~\ref{fig:lstm}.
	
	The LSTM network consists of the following components:
	\begin{itemize}
		\item \textbf{Input projection:} A fully connected layer maps the input features into the LSTM hidden layers.
		\item \textbf{LSTM layers:} Stacked LSTM layers process the projected sequence. The LSTM hidden state and cell state are initialized to zero at the first timestep.
		\item \textbf{Autoregressive output:} The network predicts the output sequence in an autoregressive manner:
		\begin{itemize}
			\item Start from the last hidden output of the LSTM after processing the input sequence.
			\item Apply a GELU activation, followed by a fully connected layer to map to the output feature dimension.
			\item Apply a $\tanh$ activation to ensure outputs are bounded in $[-1,1]$.
			\item The predicted output is fed back (via the input projection layer) into the LSTM to generate the next timestep.
		\end{itemize}
	\end{itemize}
	
	\textbf{Remark:} The autoencoder, with three convolutional layers, batch normalization, and fully connected layers, 
	has approximately $8.8 \times 10^6$ learnable parameters. 
	The subsequent LSTM, with two layers of 599 hidden units each, contains about $5.74 \times 10^6$ parameters. 
	In total, the combined model has roughly $1.45 \times 10^7$ parameters.

	\section{Idealized models}
	In this section, we discuss the idealized models, which provide pseudo-observations for the bridging model. The first model is a nonlinear stochastic PDE defined only on the equator. The second one is a six-dimensional stochastic ODE that provides SST time series in the east and central Pacific (namely, averaged SSTs over the Ni\~no 3 and Ni\~no 4 regions).
	\subsection{The simple intermediate-coupled Model} \label{sec:model}
	The starting interannual model is a deterministic and linear coupled atmosphere-ocean-SST system:
	\vspace{.4cm}
	
	\textbf{Atmosphere:} A non-dissipative Matsuno–Gill type atmosphere model \cite{gill1980,matsuno1966}:
	\begin{align}
		-y v - \partial_x \theta &= 0, \\
		y u - \partial_y \theta &= 0, \\
		-\left(\partial_x u + \partial_y v\right) &= Eq/\left(1 - \bar{Q}\right),
	\end{align}
	
	\textbf{Ocean:} A shallow-water ocean model \cite{vallis2016}:
	\begin{align}
		\partial_t U - c_1 YV + c_1 \partial_x H &= c_1 \tau_x, \\
		Y V + c_1 \partial_Y H &= 0, \\
		\partial_t H + c_1 \left(\partial_x U + \partial_Y V\right) &= 0,
	\end{align}
	
	\textbf{SST:} An SST budget equation \cite{jin1997}:
	\begin{align}
		\partial_t T = -c_1 \zeta E_q + c_1 \eta_1 H + c_1 \eta_2 U.
		\label{eq:SST}
	\end{align}

	\vspace{1em}
	\noindent where
	\(u, v\) are zonal and meridional wind speeds (atmosphere) and \(\theta\) denotes potential temperature. \(U, V\) are zonal and meridional ocean currents, \(H\) is thermocline depth, and  \(T\) represents sea surface temperature (SST). All the variables are anomalies. The coordinate variables are the interannual time coordinate $t$, the zonal coordinate $x$, while $y$ and $Y$ denote the meridional coordinates for the atmosphere and ocean components, respectively. $\bar{Q}$ is the background
	vertical moisture gradient. The latent heat is defined as $E_q = \alpha_q T$. $\zeta$ is the latent heating
	exchange coefficient, $\eta_1$ (stronger in the EP) and $\eta_2$ (stronger in the CP) are the thermocline and zonal advective feedback strengths.
	\\
	The stochastic version of the model is developed through two steps~\cite{chen2023simple}:
	
	\begin{enumerate}
		
		\item Stochastic wind bursts: adding an extra term to wind stress $\tau_x$:
		\begin{equation}
			\tau_x = \gamma(u + u_p),
		\end{equation}
		where $u_p$ is the contribution from the stochastic wind bursts with the following structure~\cite{chen2023simple}:
		\begin{equation}
			u_p(x,y,t) = a_p(t)\, s_p(x)\, \phi_0(y),
		\end{equation}
		where $\phi_0(y)$ is the leading meridional basis and $s_p(x)$ is a fixed spatial structure localized in the western Pacific (WP) due to the severity of the observed active wind bursts there. The time series $a_p(t)$ describes the wind burst amplitude and is governed by a one-dimensional real-valued stochastic process~\cite{gardiner2009stochastic}:
		\begin{equation}
			\frac{d a_p}{dt} = -d_p a_p + \sigma_p(T_C) \dot{W}_p, \tag{8}
		\end{equation}
		with $d_p$ being the damping term results in about one month of the wind's decorrelation time. $\dot{W}_p$ is a white noise source with strength $\sigma_p(T_C)$ which is state-dependent, as a function of the interannual SST~\eqref{eq:SST} averaged over the western-central Pacific. In the absence of seasonal cycle and decadal influence,
		\begin{equation}
			\sigma_p(T_C) = 1.6\left(\tanh(T_C) + 1\right).
		\end{equation}
		\textbf{Wind burst amplitude's behavior:} $a_p>0\Rightarrow$ westerly wind bursts, $a_p<0\Rightarrow$ easterly wind bursts.

		\item  {Stochastic process for the decadal variability: }
		\begin{equation}
			\frac{dI}{dt} = -\lambda (I - m) + \sigma_I(I) \dot{W}_I, \tag{9}
		\end{equation}
		where the damping $\lambda = 5~\text{years}^{-1}$ representing the decadal time scale. Similarly, $ \sigma_I(I)$ and $ \dot{W}_I$ are the state-dependent noise strength and the white noise source.

	\end{enumerate}
	Although the coupled model developed so far incorporates state-dependent noise, it remains fundamentally linear. The framework, however, falls short in capturing several key dynamical and statistical features observed in the complexity of ENSO.
	
	From a dynamical perspective, at least two significant nonlinearities need to be introduced to the model.
	\begin{enumerate}
		\item The decadal variability influences the strength of the zonal advective feedback, meaning its effect on ENSO dynamics should naturally be treated as nonlinear. Specifically, the decadal variability acts as a multiplicative factor modifying the zonal advection coefficient. Incroporating the decadal variability into the SST budget~\eqref{eq:SST}, we have:
		\begin{align}
			\partial_t T = -c_1 \zeta E_q + c_1 \eta_1 H + c_1 I \eta_2 U+{c_2}.
		\end{align}
		The small constant $c_2$ guarantees all the variables have a climatology with zero mean, regardless of the nonlinearity.
		
		\item Incorporated the damping coefficient in the SST equation. Here, $-c_1 \zeta \alpha_q$ is the damping coefficient, with $\alpha_q$ parameterized by a nonlinear quadratic function of the CP SST:
		\begin{subequations}
			\begin{align}
				\alpha_q &= q_c q_e \exp(q_e \bar{T}) / \tau_q \times \beta_1(T) \times \beta_2(t), \\
				\beta_1(T) &= 1.8 - \eta_2/3 + \left(0.2 + |T_c + 0.4| \times \eta_2\right)^2/5, \\
				\beta_2(T) &= 1 + 0.5 \sin\left(2\pi \left(t - \frac{1}{12}\right)\right) + 0.1 \sin(2\pi t) \eta_2 \nonumber \\
				&\quad - 0.0625 \sin\left(4\pi \left(t - \frac{3}{12}\right)\right) \eta_1.
			\end{align}
		\end{subequations}
		The symmetric axis of this quadratic function has a negative value, imposing a stronger damping when the CP SST is positive.
		
	\end{enumerate}
	To simplify the computation of the model solution, a meridional projection and truncation are applied to the coupled system, which is known to possess meridional basis functions in the form of parabolic cylinder functions~\cite{majda2003introduction}. For the construction of a simplified intermediate complexity model (ICM), only the leading basis function is retained for both the atmosphere and the ocean, denoted by $\phi_0(y)$ and $\psi_0(Y)$, respectively.

	\subsection{The 6-dimensional conceptual model}
	So far, we have used an intermediate-coupled model as the source of pseudo-observations. To further demonstrate the framework's skill in bridging exaggerated complexity discrepancies across the model hierarchy, we utilize a conceptual model to produce pseudo-observations.
	
	Specifically, we consider the multiscale model for El Niño complexity~\cite{chen2022multiscale}, which builds upon a deterministic three-region framework with zonal advective feedback~\cite{fang2018three} as its foundation, explicitly capturing coupled air–sea interactions across the WP, CP, and EP. The model further incorporates a stochastic representation of tropical atmospheric intraseasonal wind disturbances, such as westerly wind bursts (WWBs), easterly wind bursts (EWBs), and the Madden–Julian Oscillation (MJO). This stochastic forcing is modeled using a multiplicative noise process to reflect the modulation of wind bursts by interannual SST anomalies.
	
	In addition, the model introduces a stochastic process to represent multi-decadal variations in the background Walker circulation~\cite{yang2021enso}, which modulates both the strength and frequency of EP- and CP-type El~Niño events. Collectively, these components yield a multiscale stochastic dynamical system that integrates interannual, intraseasonal, and decadal processes. The full model reads \cite{chen2022multiscale}:
	\begin{subequations}
		\begin{align}
			\frac{du}{dt} &= -r u - \frac{\alpha_1 b_0 \mu}{2} (T_C + T_E) + \beta_u \tau + \sigma_u \dot{W}_u, \label{eq:u}\\
			\frac{dh_W}{dt} &= -r h_W - \frac{\alpha_2 b_0 \mu}{2} (T_C + T_E) + \beta_h \tau + \sigma_h \dot{W}_h, \label{eq:hW}\\
			\frac{dT_C}{dt} &= \left(\frac{\gamma b_0 \mu}{2}  - c_1(T_C) \right) T_C + \frac{\gamma b_0 \mu}{2} T_E + \gamma h_W + \sigma_u + C_u + \beta_C \tau + \sigma_C \dot{W}_C, \label{eq:TC}\\
			\frac{dT_E}{dt} &=\gamma h_W + \left( \frac{3\gamma b_0 \mu}{2}  - c_2 \right) T_E - \frac{\gamma b_0 \mu}{2} T_C + \beta_E \tau + \sigma_E \dot{W}_E, \label{eq:TE}\\
			\frac{d\tau}{dt} &= -d_\tau \tau + \sigma_\tau(T_C) \dot{W}_\tau, \label{eq:tau}\\
			\frac{dI}{dt} &= -\lambda(I - m) + \sigma_I(I) \dot{W}_I. \label{eq:I}
		\end{align}
	\end{subequations}
	Here, the interannual component (Eqs.~\ref{eq:u}--\ref{eq:TE}) governs the deterministic dynamics for both CP- and EP-type ENSO events. The intraseasonal component (Eq.~\ref{eq:tau}) represents random wind bursts, while the decadal component (Eq.~\ref{eq:I}) captures slow variations in the background strength of the Pacific Walker circulation. \(T_C\) and \(T_E\) are SST anomalies in the central and eastern Pacific, respectively. \(u\) denotes the ocean zonal current anomaly in the CP.  \(h_W\) is the thermocline depth anomaly in the WP. \(\tau\) is the amplitude of intraseasonal random wind bursts, including MJO.  \(I\) is an index representing the background Walker circulation, equivalent to the zonal SST difference between WP and CP, which controls the strength of zonal advective feedback. Next, the bivariate regression method allows for the reconstruction of the spatiotemporal evolution of the SST field:
	\begin{equation}
		\mathrm{SST}(x,t) = r_C(x) \, T_C(t) + r_E(x) \, T_E(t),
		\label{eq:sst_regression}
	\end{equation}
	where \(x\) denotes longitude and \(t\) represents time. The regression coefficients \(r_C(x)\) and \(r_E(x)\) are estimated from observational data at each longitude grid point. The temporal components \(T_C(t)\) and \(T_E(t)\), corresponding to the Niño3 and Niño4 indices obtained from the model, are then substituted into Eq.~\eqref{eq:sst_regression} to yield the SST spatiotemporal patterns.

	\section{EnKF implementation}
	
\begin{figure}[htbp]
		\centering
		\includegraphics[width=.7\textwidth]{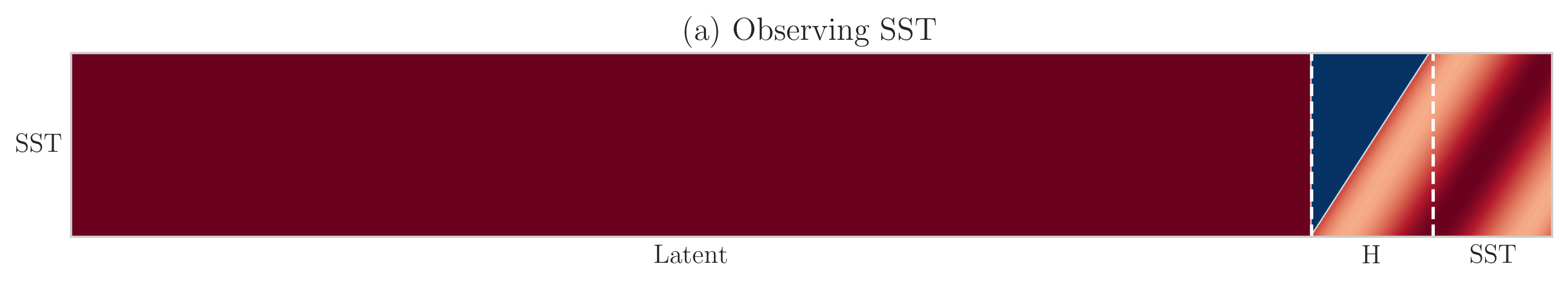}
		\includegraphics[width=.7\textwidth]{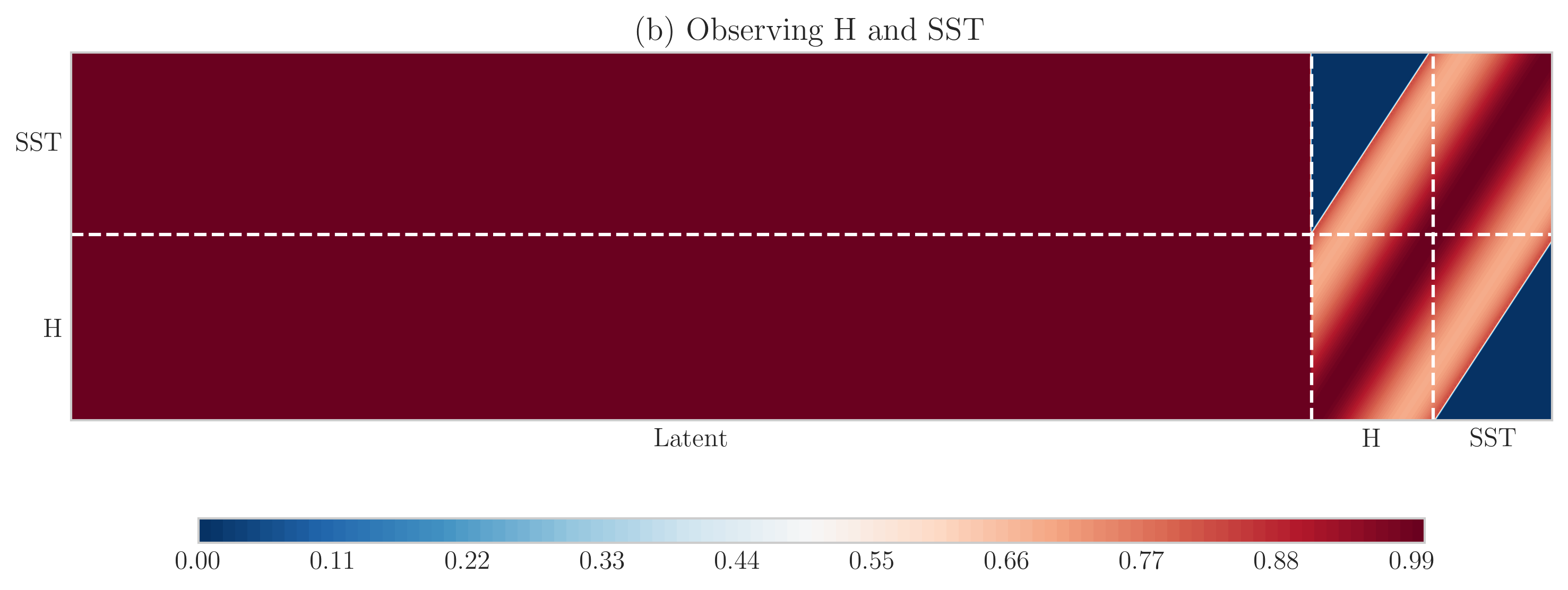}
		\caption{The localization matrix $L$.}\label{fig:local}
	\end{figure}

	We modify the basic EnKF implementation to improve numerical stability, prevent sampling collapse, and enable localized updates in the mixed latent/physical state. The main additions are listed below:
	
	\begin{enumerate}
		\item \textbf{Multiplicative inflation.}
		Since we trust the pseudo-observations with high confidence, the analysis step may reduce ensemble spread. Therefore, we apply multiplicative inflation to the ensemble anomalies immediately before covariance computation:
		\begin{equation}
			\mathbf{x}^{f,(i)}_{t+1} \leftarrow (1-\alpha) \sum_{j=1}^N \mathbf{x}^{f,(j)}_{t+1} + \alpha\,\mathbf{x}^{f,(i)}_{t+1},
		\end{equation}
		where $\alpha$ is the inflation factor (here, $\alpha=1.09$). Inflation keeps the forecast covariance sufficiently large in the observed directions, ensuring that the Kalman gain is not a numerical challenge.
		
		\item \textbf{Robust inversion and adaptive regularization of the innovation covariance.}
		The innovation covariance
		\begin{equation}
			\mathbf{S} = \mathbf{H} \mathbf{P}_{xb} \mathbf{H}^\top + \mathbf{R},
		\end{equation}
		where $\mathbf{P}_{xb}$ is the forecast error covariance with starting value of identity matrix, $\mathbf{R}$ is the observation noise covariance, set as 4\% of observation energy, 
		and $\mathbf{H}$ is the observation operator. $\mathbf{S}$ is inverted using an SVD-based pseudo-inverse with relative cutoff \(\tau = 10^6\cdot\sigma_{\max}(\mathbf{S})\). The condition number \(\kappa(\mathbf{S})\) exceeding a  threshold, a small nugget \(\delta I\) is added with
		\begin{equation}
			\delta=\max\left(10^{-5}\cdot\sigma_{\max}(\mathbf{S}),\ 10^{-10}\right)
		\end{equation}
		before inversion. $\sigma_{\max}(S)$ denotes the maximum eigenvalue of $S$. This prevents amplification of numerical noise due to tiny singular values.
		
		\item \textbf{Kalman gain capping.}
		As an additional safety mechanism, the computed Kalman gain
		\begin{equation}
			\mathbf{K} = \mathbf{P}_{xz}^{\text{loc}}\,\mathbf{S}^{-1}
		\end{equation}
		is scaled to enforce $\|\mathbf{K}\| \le \mathbf{K}_{\max}$ ($\mathbf{K}_{\max} = 10^4$). This avoids explosive analysis increments. $\mathbf{P}_{xz}^{\text{loc}}$ denotes the localized cross covariance. 
		
		\item \textbf{Schur-product localization (Gaspari--Cohn) .}
		
		We apply a localization technique, the Gaspari-Cohn localization matrix, which filters out weak correlations beyond a certain distance and allows us to capture the spatial dependency among the physical variables in the augmented space. The Gaspari--Cohn localization function \cite{gaspari1999construction} is defined as
		\begin{equation}
			\rho(r) =
			\begin{cases}
				1 - \frac{5}{3}\left(\frac{r}{c}\right)^2
				+ \frac{5}{8}\left(\frac{r}{c}\right)^3
				+ \frac{1}{2}\left(\frac{r}{c}\right)^4
				- \frac{1}{4}\left(\frac{r}{c}\right)^5,
				& 0 \leq \frac{r}{c} \leq 1, \\[1em]
				-\frac{2}{3}\left(\frac{r}{c}\right)^{-1}
				+ 4 - 5\left(\frac{r}{c}\right)
				+ \frac{5}{3}\left(\frac{r}{c}\right)^2
				+ \frac{5}{8}\left(\frac{r}{c}\right)^3
				- \frac{1}{2}\left(\frac{r}{c}\right)^4
				+ \frac{1}{12}\left(\frac{r}{c}\right)^5,
				& 1 < \frac{r}{c} \leq 2, \\[1em]
				0, & \frac{r}{c} > 2,
			\end{cases}
		\end{equation}
		where $r$ is the distance between two state variables and $c$ is the localization radius. Therefore, we apply the covariance localization as
		\begin{equation}
			P_{xz}^{\text{loc}} = P_{xz}\circ L,
		\end{equation}
		where \(\circ\) denotes the Hadamard product.

		The localization matrix \(L \in \mathbb{R}^{n_l \times n_o}\) regulates the influence between the latent and physical variables in the cross-covariance computation. It is precomputed once using the Gaspari--Cohn localization function and remains constant throughout the data assimilation process.
		
		In particular, we build \(L\) by evaluating pairwise distances between state and observation positions, normalized by the length of each observation variable $\ell_{\text{obs i}}$ (e.g., we observe physical variables at 49 grid points and set $\ell_{\text{obs i}}=49$.). For the physical part of the state vector, these normalized distances are used as inputs to the Gaspari--Cohn tapering function. For the latent part of the state, no localization is applied, and $L_{ij} = 1$ to preserve the full coupling in the latent dimensions.
		\[
		L_{ij} =
		\begin{cases}
			1, & \text{for latent rows,}\\[8pt]
			\rho\!\left(\dfrac{|x_i - x_j|}{\ell_{\text{obs i}}}\right), & \text{for physical rows.}
		\end{cases}
		\]
	\end{enumerate}
	We show the matrix $L$ in figure~\ref{fig:local} for two cases where we observe only SST and SST-H.
	
	\section{Additional results for CESM model}
\begin{figure}[htbp]
		\centering
		\includegraphics[width=\textwidth]{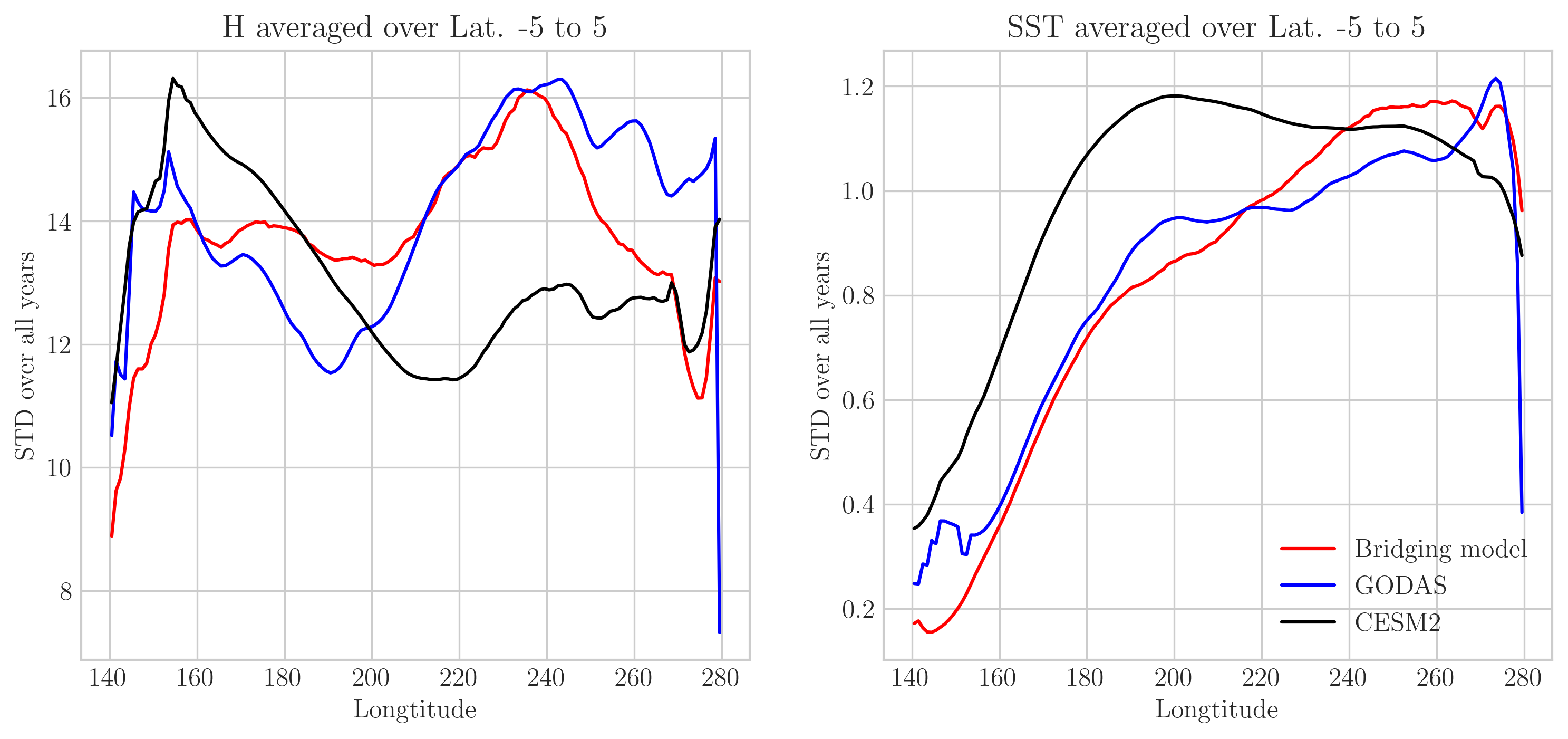}
		\caption{The standard deviation of the thermocline depth and SST averaged from latitudes ($5^\circ$S-$5^\circ$N). Comparing the GODAS, CESM2, and bridging model.}\label{fig:CESM_avg}
	\end{figure}

\begin{figure}[htbp]
		\centering
		\includegraphics[width=.7\textwidth]{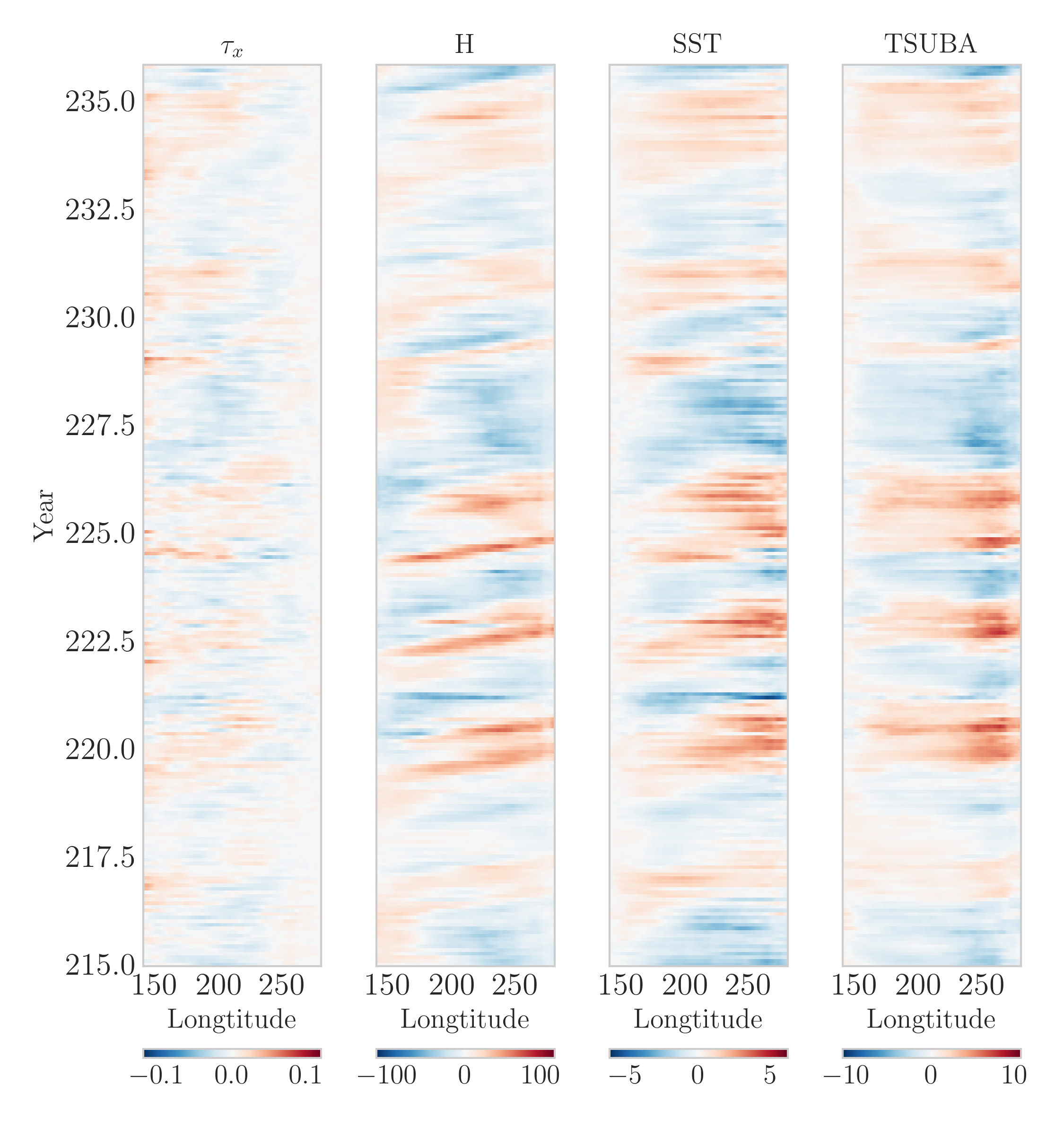}
		\caption{Hovm\"oller diagram of variables from the bridging model that is built based on the CESM model.}\label{fig:2Var_Hov_5}
	\end{figure}

	Figure \ref{fig:CESM_avg} compares the SST and the thermocline depth of CESM2, the bridging model, and the reanalysis GODAS averaged from latitudes ($5^\circ$S-$5^\circ$N). The results presented here are consistent with those shown in the main text (for SST along the equator), which demonstrate the skill of the bridging model in capturing SST variability.
	Figure \ref{fig:2Var_Hov_5} depicts a 20-year period of spatiotemporal patterns for zonal wind stress ($\tau_x$), thermocline depth (H), subsurface temperature (TSUBA), and sea surface temperature (SST). It reveals the coupled ocean-atmosphere dynamics of ENSO. An El Ni\~no event often initiates with a westerly wind burst in the western Pacific, corresponding to positive $\tau_x$ anomalies. These winds force oceanic Kelvin waves that propagate eastward, depressing the thermocline (positive H anomalies) and warming the subsurface (positive TSUBA anomalies). As the warm water reaches the surface, it triggers a positive SST anomaly, establishing a feedback loop. Here, the SST anomaly weakens the east-west pressure gradient, allowing the anomaly to grow and propagate eastward.
	
	The diversity of ENSO is determined by the precise interplay of these fields. If the wind forcing is concentrated in the central Pacific, the warming remains localized, leading to a CP El Ni\~no. Conversely, if the coupled feedback drives the anomalies far to the east, an EP El Ni\~no develops, often with greater intensity due to the larger background SSTs. The TSUBA acts as a reservoir of heat, with its anomalies leading those at the surface and providing the oceanic memory for the cycle. The subsequent transition to the La Ni\~na phase corresponds to the recharge of the heat content and the shutdown of the warming.
	
	\section{Bridging model results based on the GFDL CM4 operational model}
	
\begin{figure}[htbp]
		\centering
		\includegraphics[width=1\textwidth]{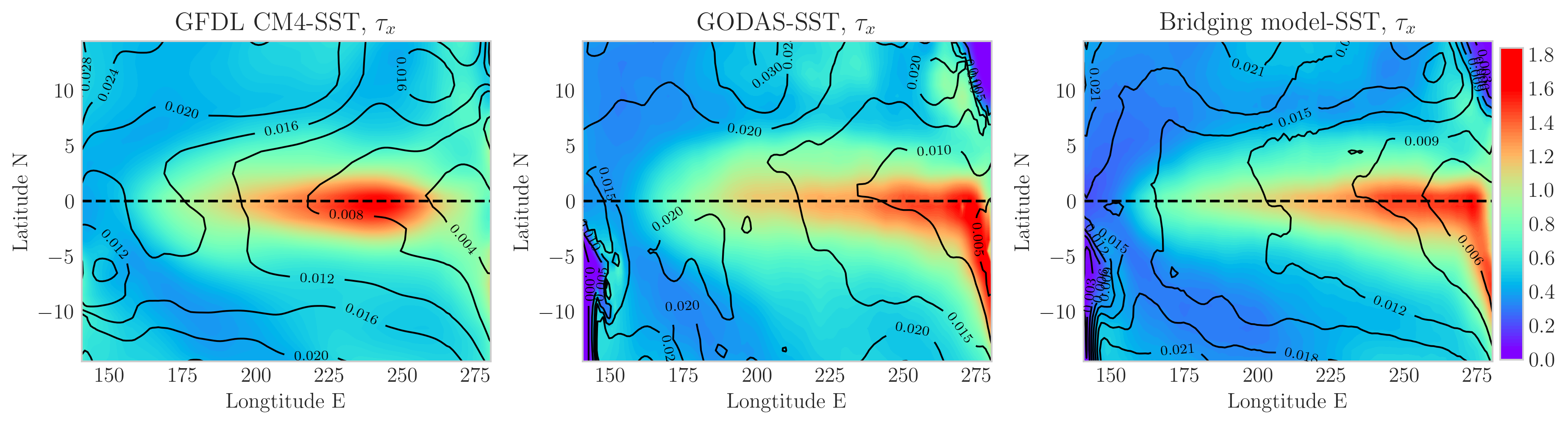}
		\includegraphics[width=\textwidth]{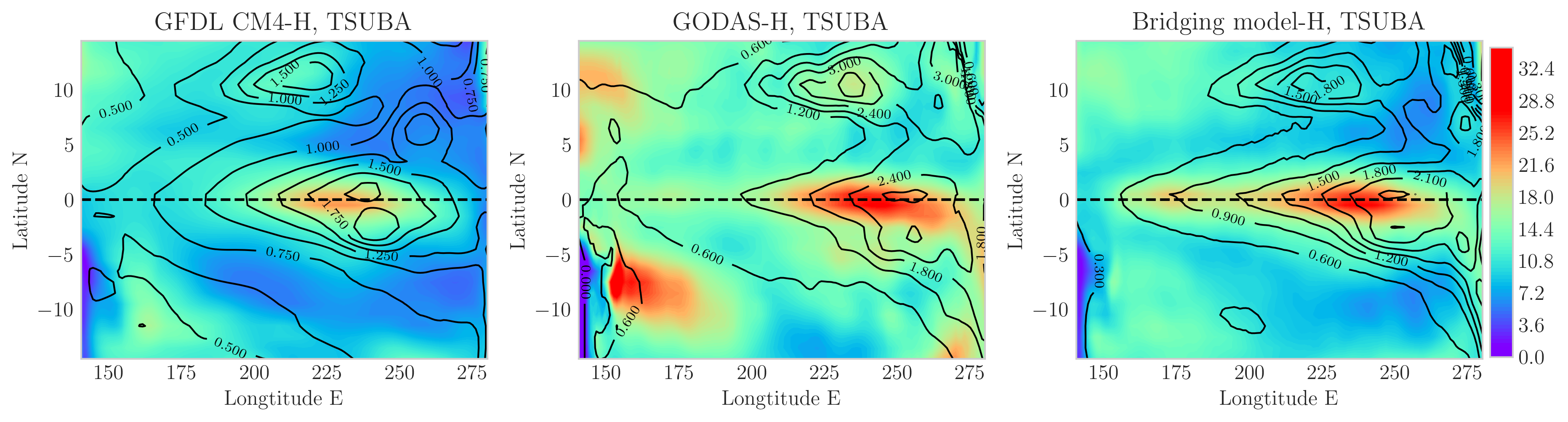}
		\caption{Comparison of interannual variability patterns, measured by the standard deviation, among GFDL CM4, GODAS reanalysis, and the bridging model. The left, middle, and right columns show results from GFDL CM4, GODAS, and the bridging model, respectively. The top row displays sea surface temperature (SST, shading) and zonal wind stress ($\tau_x$, contours). The bottom row shows thermocline depth ($H$, shading) and subsurface temperature (TSUBA, contours). The idealized used here is the CF23 model.}
		\label{fig:GFDL_con}
	\end{figure}

	In this section, we briefly report the resulting bridging model, which utilizes the dataset obtained from the GFDL CM4 model as training data for the LSTM forecast model in the latent space. The idealized model that generates pseudo-observations is still the CF23 intermediate-coupled model (Figures \ref{fig:GFDL_con}--\ref{fig:GFDL_equator}) and the CF22 conceptual model (Figure \ref{fig:concept_SST}). The goal is to validate the robustness of our framework. We follow the autoencoder and augmented structures discussed in previous sections. It is seen that the bias from the SST and thermocline depth in the GFDL CM4 is almost eliminated in the bridging model. The GFDL CM4 has a more skillful characterization of the off-equator behavior, which the bridging model inherits. The bridging model remains adept at generating various types of ENSO events and captures the statistical features of ENSO diversity and complexity.
	
	\begin{figure}[htbp]
		\centering
		\includegraphics[width=1\textwidth]{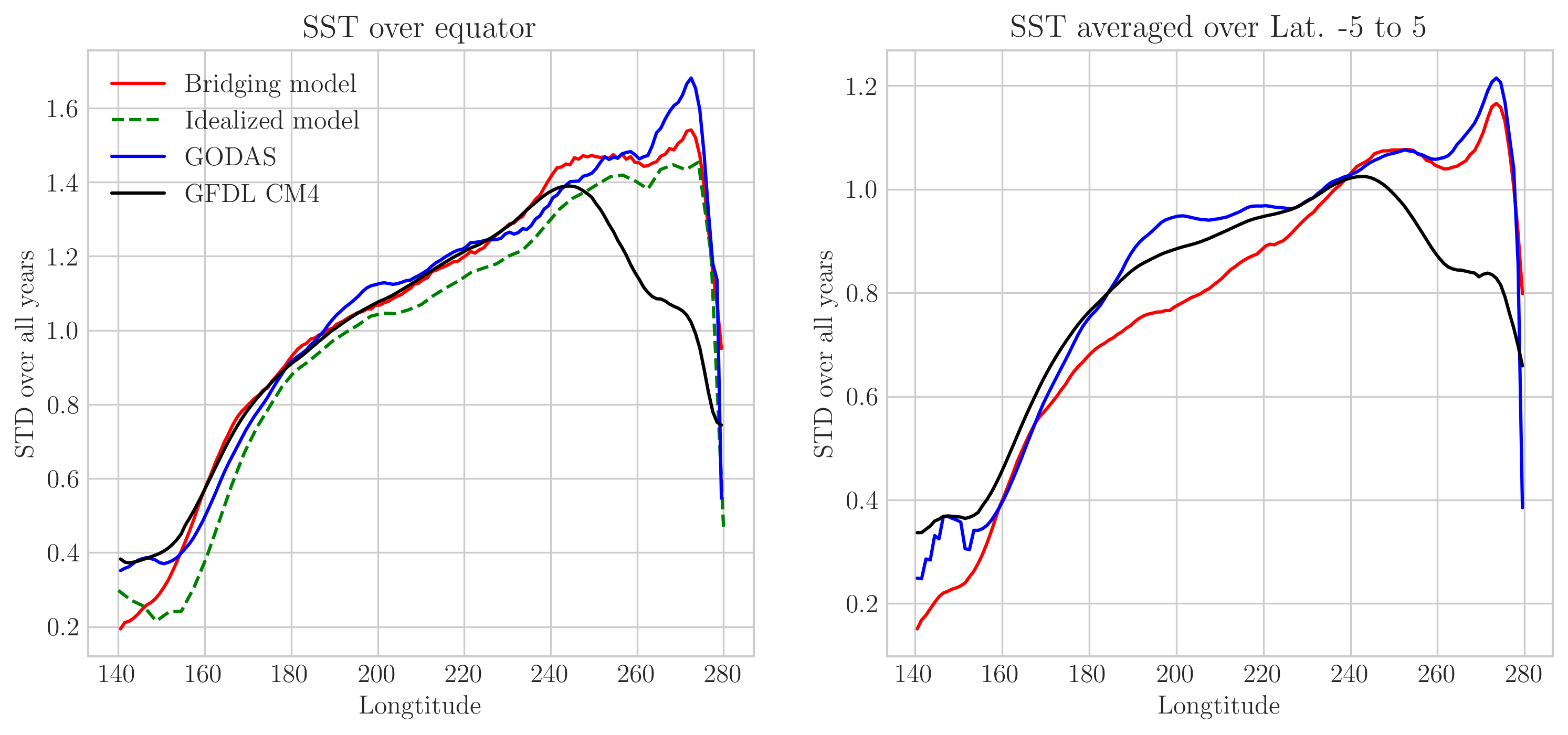}
		
		\includegraphics[width=\textwidth]{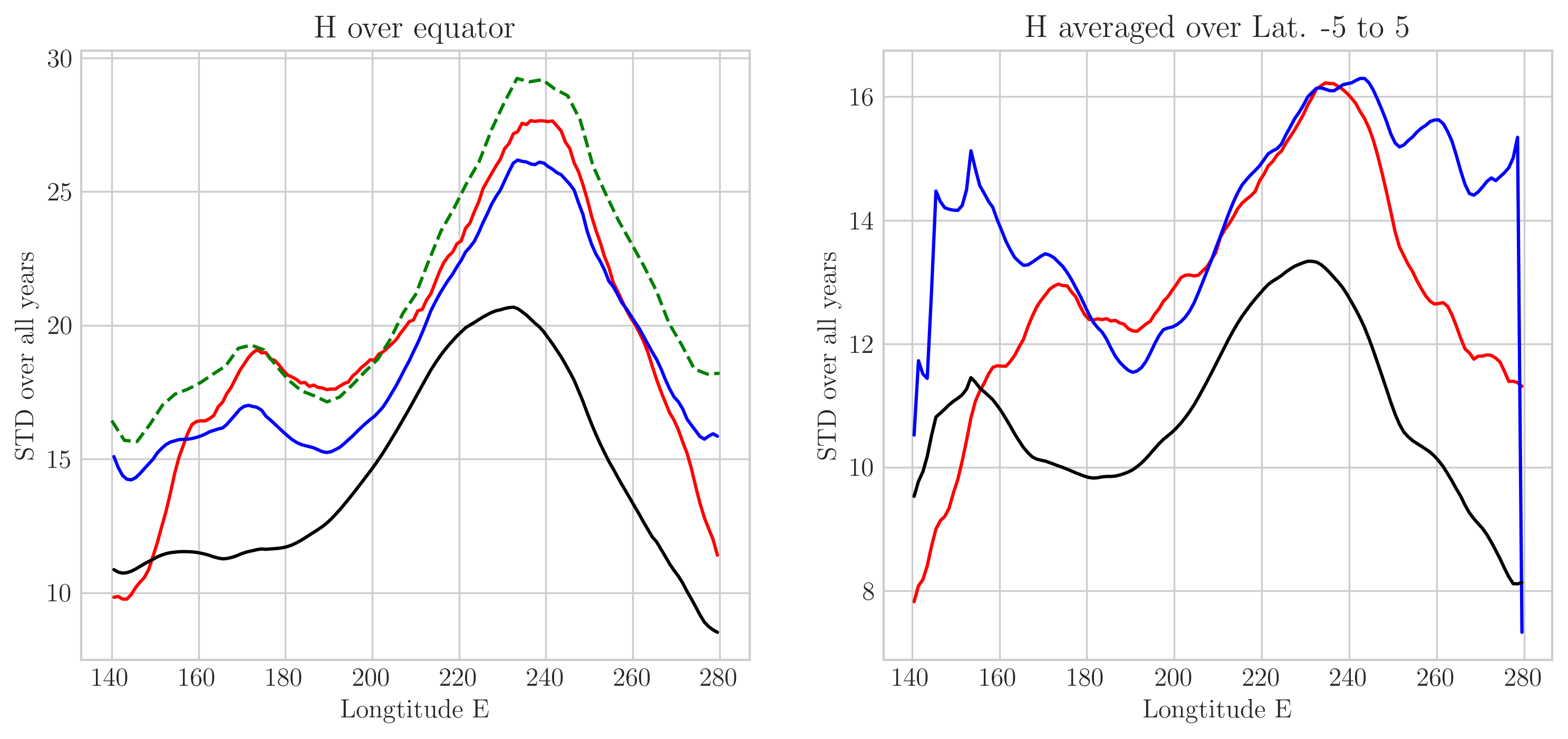}
		\caption{The standard deviation of SST (top) along the equator and averaged from latitudes ($5^\circ$S-$5^\circ$N) and the thermocline depth (bottom). Comparing the GODAS, GFDL CM4, and bridging model, where the idealized model used here is the CF23 model. }
		\label{fig:GFDL_equator}
	\end{figure}

	\begin{figure}[htbp]
		\centering
		\includegraphics[width=\textwidth]{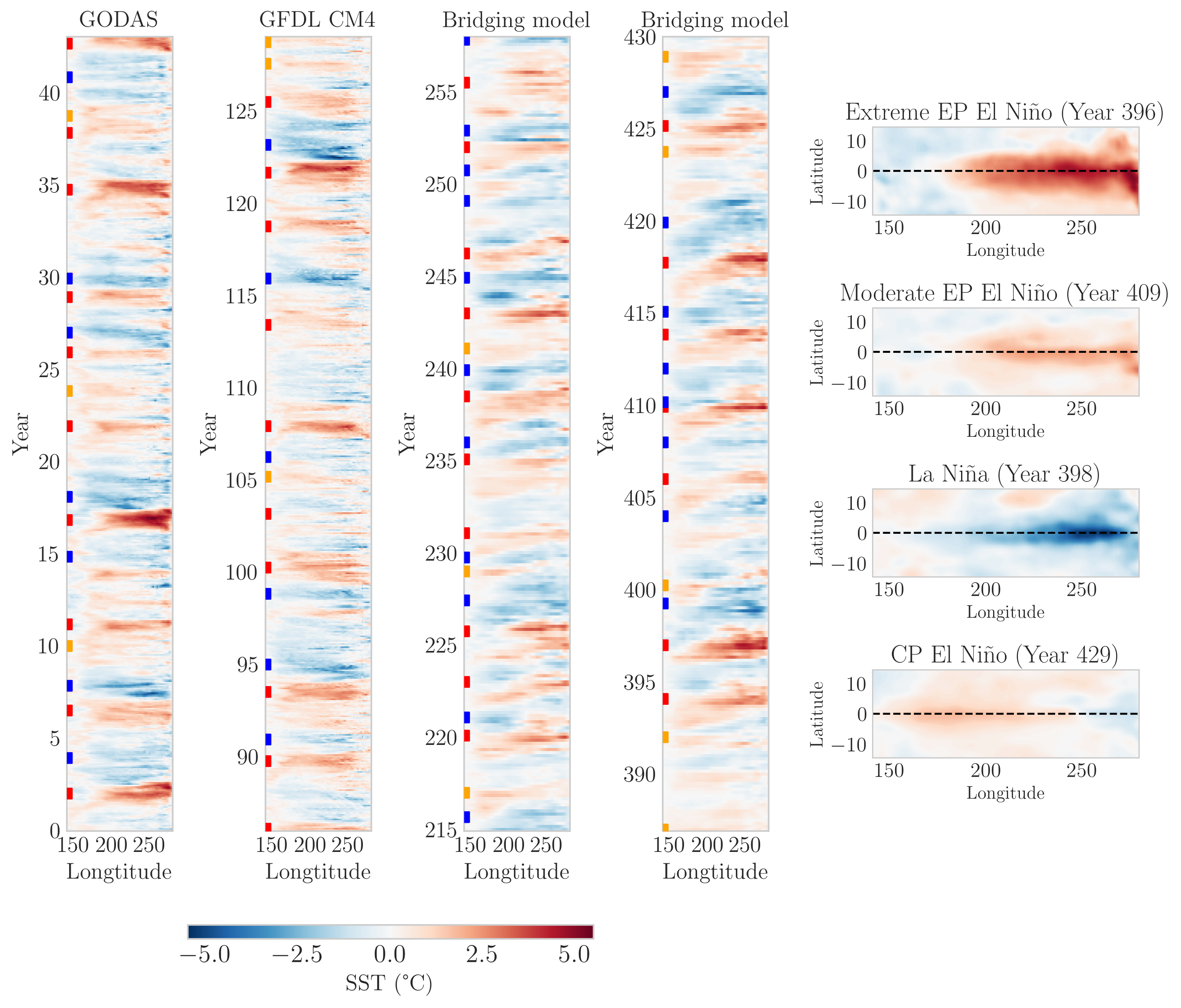}
		\caption{Hovm\"oller diagram of SST and the captured events using different models. The bridging model used the CF23 model.}
		\label{fig:GFDL_event}
	\end{figure}
	
	\begin{figure}[htbp]
		\centering
		\includegraphics[width=.7\textwidth]{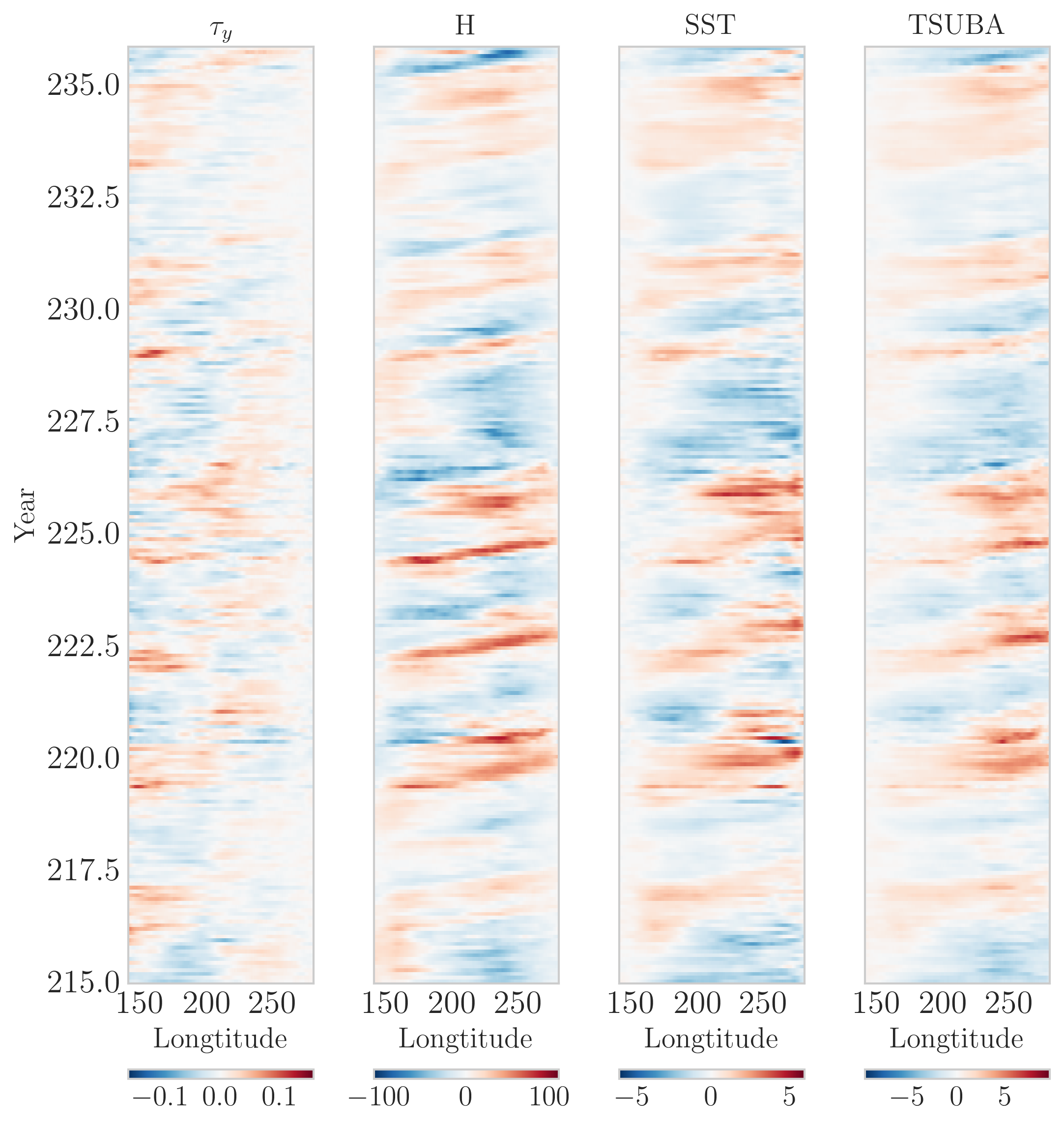}
		\caption{Hovm\"oller diagram of different fields from the bridging model, where the idealized used here is the CF23 model.}
		\label{fig:GFDL_equator}
	\end{figure}
	
	\begin{figure}[htbp]
		\centering
		\includegraphics[width=\textwidth]{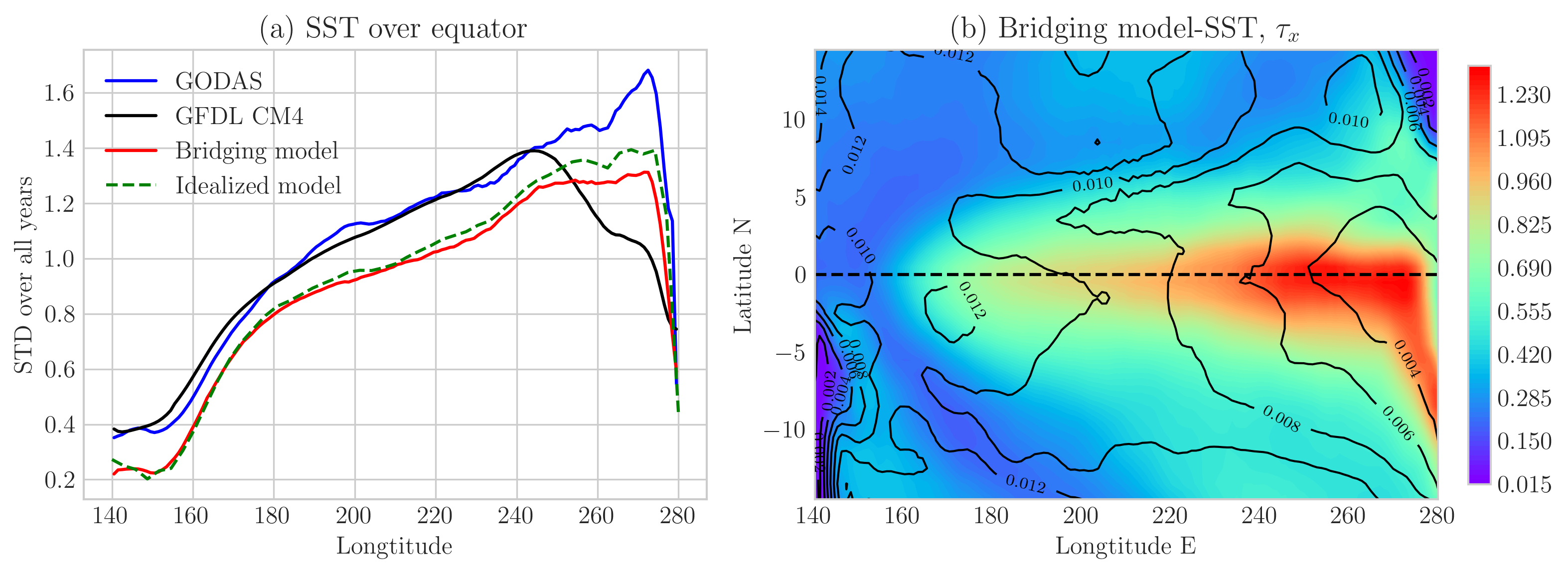}
		\caption{Comparison of equatorial Pacific SST, measured by the standard deviation,
			among GFDL CM4, GODAS reanalysis, and the bridging model with pseudo-observations provided by even a simpler 6-dimensional stochastic ODE model (CFY22 model). The right
			displays (SST, shading) and zonal wind stress ($\tau_x$, contours).  }
		\label{fig:concept_SST}
	\end{figure}

\end{document}